\documentclass[sn-basic]{sn-jnl}

\usepackage[utf8]{inputenc}
\usepackage{graphicx}

\usepackage{amsthm}
\usepackage[fleqn]{amsmath}
\usepackage{amssymb}
\usepackage{siunitx}
\usepackage{xspace}
\usepackage{subcaption}
\usepackage{calrsfs}
\usepackage[fleqn]{mathtools}
\usepackage{color}
\usepackage{booktabs}
\usepackage{setspace}
\usepackage{float}
\usepackage{hyperref}
\usepackage{algorithm}
\usepackage[noend]{algpseudocode}
\usepackage{listings}
\usepackage[htt]{hyphenat}
\usepackage[autostyle, english = american]{csquotes}
\usepackage{url}
\usepackage{nomencl}
\usepackage{multirow}

\usepackage{mathrsfs}%
\usepackage[title]{appendix}%
\usepackage{textcomp}%
\usepackage{manyfoot}%
\usepackage{booktabs}%
\usepackage{algorithm}%
\usepackage{algorithmicx}%
\usepackage{algpseudocode}%
\usepackage{listings}%
\usepackage{tabularx}

\DeclareMathAlphabet{\pazocal}{OMS}{zplm}{m}{n}
\DeclareMathOperator*{\argmax}{argmax}

\newcommand{\stt}[1]{{\small\texttt{#1}}}
\newtheorem{defn}{Definition} 
\newcommand{\vect}[1]{\mathbf{#1}}

\makenomenclature

\begin{document}

\title[From Data to Robot Action with Expert-Guided Refinement]{Inductive Learning of Robot Task Knowledge from Raw Data and Online Expert Feedback}


\author*[1]{\fnm{Daniele} \sur{Meli}}\email{daniele.meli@univr.it}

\author[2]{\fnm{Paolo} \sur{Fiorini}}

\affil*[1]{\orgdiv{Department of Computer Science}}

\affil[2]{\orgdiv{Department of Engineering for Innovation Medicine}, \orgname{University of Verona}, \orgaddress{\street{Strada Le Grazie 15}, \city{Verona}, \postcode{37135}, \state{Italy}}}


\abstract{
The increasing level of autonomy of robots poses challenges of trust and social acceptance, especially in human-robot interaction scenarios.
This requires an interpretable implementation of robotic cognitive capabilities, possibly based on formal methods as logics for the definition of task specifications. However, prior knowledge is often unavailable in complex realistic scenarios.

In this paper, we propose an \emph{offline algorithm} based on inductive logic programming from noisy examples to extract task specifications (i.e., action preconditions, constraints and effects) directly from raw data of few heterogeneous (i.e., not repetitive) robotic executions. Our algorithm leverages on the output of \emph{any} unsupervised action identification algorithm from video-kinematic recordings. Combining it with the definition of very basic, almost task-agnostic, commonsense concepts about the environment, which contribute to the interpretability of our methodology, we are able to learn logical axioms encoding preconditions of actions, as well as their effects in the event calculus paradigm. 

Since the quality of learned specifications depends mainly on the accuracy of the action identification algorithm, we also propose an \emph{online framework} for incremental refinement of task knowledge from user's feedback, guaranteeing safe execution.

Results in a standard manipulation task and benchmark for user training in the safety-critical surgical robotic scenario, show the robustness, data- and time-efficiency of our methodology, with promising results towards the scalability in more complex domains.
}

\keywords{Inductive Logic Programming, Answer Set Programming, Learning Under Uncertainty, Online Learning, Human-Robot Interaction, Explainable AI}



\maketitle

\section{Introduction}
Robots are becoming increasingly popular in many fields of industry and society.
As their degree of autonomy, hence \emph{intelligence}, increases, human-robot interaction is a crucial aspect of design in the development of robotic systems, to preserve human safety and increase social acceptance in complex, typically human-centered scenarios.
In this context, recent international provisions and regulations, e.g., by the European Union\footnote{Artificial Intelligence Act, \href{https://www.europarl.europa.eu/doceo/document/TA-9-2023-0236_EN.html}{https://www.europarl.europa.eu/doceo/document/TA-9-2023-0236\_EN.html}} prescribe \emph{interpretability} \citep{rosenfeld2019explainability} as one fundamental requirement to achieve safe and transparent human-robot interaction.
To this aim, formal representation of agents' (robots') environments and tasks is a suitable and convenient choice.
Such representations in robotics include, among others, action languages for planning \citep{erdem2012applications}, as the well-established Planning Domain Definition Language (PDDL) by \cite{haslum2019introduction, estivill2013path, kootbally2015towards}, logic programming for plan re-configuration \citep{meli2023logic}, ontology-based systems \citep{manzoor2021ontology} and behavior trees \citep{ogren2022behavior}.
One fundamental limitation of these approaches is the assumption of full scenario and task knowledge, which is clearly impractical in complex, uncertain or partially observable domains.

In this paper, we propose a methodology to derive logical task specifications from visual and kinematic data of robotic executions. In particular, we assume prior knowledge of very basic commonsense concepts (\emph{features}) about the environment, e.g., related to relative positions of objects in the scene. These are usually shared among many tasks and scenarios, and they are already encoded in available general-purpose ontologies for robotic and automation, e.g., by \cite{schlenoff2012ieee}. We then combine automatic retrieval of environmental features from input execution data, with unsupervised action identification. The resulting matching examples of action execution and environmental features are used in an Inductive Logic Programming (ILP) by \cite{muggleton1994inductive} framework to derive salient task specifications, namely, preconditions and effects of actions and constraints.
Learned specifications are affected by the noise in the input data and their post-processing (i.e., environmental features retrieval and unsupervised action identification), thus they may not immediately applied for safe fully autonomous robotic planning and execution.
Hence, we finally propose a framework interleaving human and autonomous execution, allowing for incremental knowledge refinement via robot teaching.

This paper then makes the following contributions:
\begin{itemize}
    \item we propose ILP under the semantics of Answer Set Programming (ASP) by \cite{lifschitz2019answer} to learn task specifications from little noisy raw data of robotic execution. Our methodology can work in combination with any action identification algorithm, exploiting the output of the latter to define noise in action examples, and is robust . Moreover, following our previous work in \cite{meli2021inductive}, we employ the representation of event calculus \citep{mueller2008event} to efficiently learn temporal specifications about environmental features, i.e., effects of actions;
    \item we propose an ILP- and ASP-based framework for time- and data-efficient incremental learning of task knowledge and autonomous robotic planning, involving an expert human in the loop to guarantee safe execution. The proposed framework represents an extension of our preliminary work in \cite{meli_intra}, where we did not learn effects of actions and showcased only a simple example of precondition refinement for one action, without evaluating the performance quantitatively for the full task;
    \item we evaluate our approach with different action identification algorithms and with a partial ablation study (considering or neglecting \emph{unobserved action examples}). We operate in the domain of surgical task automation, specifically in the benchmark peg transfer scenario from the Fundamentals of Laparoscopic Surgery (FLS) in \cite{FLS}, using da Vinci robot and Research Kit (dVRK) \citep{kazanzides2014open}. 
    Broadly speaking, the surgical scenario poses fundamental challenges in the field of task knowledge retrieval, e.g., the scarce availability of data due to legal and organisational issues, and the paramount importance of safety and social acceptance, thus justifying the use of data-efficient and interpretable learning with ILP. With particular reference to the peg transfer scenario, it is one of the most challenging tasks from FLS, which must be accomplished in the smallest time and with the highest number of successful repetitions\footnote{\href{https://www.flsprogram.org/technical-skills-training-curriculum/}{https://www.flsprogram.org/technical-skills-training-curriculum/}}. Indeed, the peg transfer requires dual arm coordination and reasoning on the placements of rings and pegs, which stresses the mental focus and resistance of the surgeon under time constraints in a real operating room. Moreover, this task is not only paradigmatic in the clinical context, but it is also an example of a generic structured robotic pick-and-place task. Thus, our experiments are designed to show promising results towards the general applicability of our methodology to more general robotic and automation domains;
    \item we make publicly available the code for replicating experimental results and the framework for incremental task knowledge refinement\footnote{\href{https://github.com/danm11694/ILASP_robotics.git}{https://github.com/danm11694/ILASP\_robotics.git}}.
\end{itemize}
\noindent
In the following, we first revise state of the art in robotic task learning, in Section \ref{sec:sota}. Then, in Section \ref{sec:background} we provide relevant background about the fundamental components of our framework, ASP and ILP, as well as a description of the peg transfer task useful as a leading example. Finally, we detail our methodology in Section \ref{sec:met}, perform experimental validation in Section \ref{sec:res} and draw conclusions with limitations and future perspectives in Sections \ref{sec:disc}-\ref{sec:conc}.

\section{Related Works}\label{sec:sota}
Learning robotic task sequences from previous observations or human teaching is a widely studied problem, especially in the area of industrial assembly and manipulation \citep{zhu2018robot,mukherjee2022survey}.
State-of-the-art approaches rely on probabilistic models to capture the variability of the environment and execution, e.g., Gaussian Mixture Models (GMMs) as in \cite{calinon2016tutorial,10.1007/s10514-018-9725-6} and Hidden Markov Models (HMMs) as in \cite{tanwani2016learning,medina2017learning}. 

With the recent uptake of deep learning and large data models, neural architectures have been employed both for online and offline learning of optimal robotic strategies for complex domains. Main approaches involve reward shaping to learn efficient task policies from human demonstrations \citep{alakuijala2023learning}, or end-to-end learning of sensor-action maps for robotic navigation \citep{liu2017learning} and manipulation \citep{rahmatizadeh2018vision}.
This has paved the way towards \emph{human-in-the-loop} Deep Reinforcement Learning (DRL), where humans communicate with the autonomous robot either to show it safe and best practice under different situations \citep{huang2024safety}, acting as teachers \citep{shenfeld2023tgrl}, or to give targets and commands, based on pre-trained foundational visual-language models \citep{shah2023lm}.

In the specific field of robotic surgery, HMMs have been used, e.g., by \cite{berthet2016hubot}, though more general Bayesian networks are the most popular \citep{blum2008modeling,charriere2017real}. Recently, DRL is becoming a prominent approach to surgical task automation \citep{shahkoo2023autonomous,fan2024learn,corsi2023constrained}, paving the way towards interactive human-robot learning \citep{long2023human}.

One crucial limitation, especially for DRL and deep learning methods, is their scarce interpretability and the lack of safety guarantees, even when safe \citep{shahkoo2023autonomous} or human-in-the-loop \citep{huang2024safety} solutions are employed. Indeed, these approaches heavily rely on large and diverse training datasets, failing at the generalization to similar but different task situations. For this reason, some researchers have attempted to directly learn human models of agency \citep{nikolaidis2015efficient,chen2020trust}, which however severely limit the performance of the autonomous agent.
Similar limitations hold for Bayeesian and statistical methods.
As an example in the surgical setting, 80 homogeneous (i.e., under the same environmental conditions) executions are required by \cite{berthet2016hubot} to learn a HMM for a good replication of the yet relatively simple peg transfer task.
Finally, these approaches are highly sensitive to parameter initialization, as shown by \cite{medina2017learning}.

In contrast, in this paper we propose to use ILP \citep{Mugg91} to learn \emph{task specifications}, i.e., preconditions, effects and constraints on actions; and then perform automated reasoning in on specifications in ASP formalism for task planning and execution.
Originated from the broad field of domain knowledge discovery \citep{mohan:ACS18,mota:aaaisymp20,laird:IS17}, ILP solves many limitations of above machine learning and DRL approaches, being data- and time-efficient, and able to easily generalize theory from data. Furthermore, the interpretable logical formalism allows easier interaction with humans for online task knowledge refinement.
For this reason, ILP has been successfully applied to robotic learning \citep{CM19,meli2020towards,meli2021inductive}.

However, one important limitation of ILP, and in general frameworks for high-level knowledge extraction, is the poor performance in the presence of \emph{noisy datasets}, i.e., task examples with uncertainty.
For this reason, researchers have tried to embed uncertainty into the inductive process, as \cite{katzouris2023online} to learn event definitions and \cite{law2023conflict} under the more generic answer set semantics.
In this direction, neurosymbolic approaches have been proposed, e.g., differentiable inductive logic programming \citep{evans2018learning,shindo2021differentiable} and statistical relational learning \citep{marra2024statistical,hazra2023deep}, which however suffer from scalability issues and cannot be currently applied out of toy robotic domains, unless strict task-specific constraints are imposed to the ILP problem \citep{hazra2023deep}. For instance, we tried to apply the publicly available code released by \cite{evans2018learning} to the peg transfer domain, but it failed for the large space of possible specifications to be learned.

Under the framework of ILP in the ASP semantics, we stem from our previous works on fully supervised task specifications learning in \cite{meli2020towards,meli2021inductive}, considering the more realistic use case of ILP examples generated from an unsupervised action identification algorithm applied on raw robotic execution data. 
We then extend \cite{meli_intra} to refine task knowledge (including effects of actions) from human feedback via online ILP, guaranteeing safe execution thanks to expert supervision. 
We use the state-of-the-art ILASP (Inductive Learning of Answer Set Programs) tool by \cite{law2018inductive} for its theoretical and empirical advantages against main competitors as \cite{katzouris2016online,schuller2018best,cropper2021learning}, encompassing learning from noisy data, context-dependent examples (capturing the action-environment connection particularly relevant in robotic tasks) and the possibility to induce \emph{most probable} ASP theories, i.e., covering most of the examples.
Moreover, the answer set semantics supported by ILASP is the state of the art for planning \citep{erdem2016applications}, especially in robotic contexts \citep{erdem:KI18,meli2023logic}.

\section{Background and notation}\label{sec:background}
In this section, we describe the paradigmatic peg transfer task, to use it as a leading example in the rest of the paper. We then provide basic notions about ASP and ILP under the ASP semantics, necessary to detail our methodology next.

We start by providing a list of relevant symbols used throughout the paper, as a practical reference to the reader.

\subsection{Notation}
\begin{tabularx}{\textwidth}{lX}
\stt{text} & Atoms in ASP formalism\\
\stt{:-} & Logical implication $\leftarrow$ in ASP syntax\\
$\models$ & Logical entailment\\
$\pazocal{H(a)}$ & Herbrand base of $a$, representing the set of all possible ground versions of $a$\\
$B$ & Background knowledge for ILP task\\
$S_M$ & Search space for ILP task\\
$E = \langle E^+, E^- \rangle$ & Sets of positive ($E^+$) and negative ($E^-$) examples for ILP task\\
$H$ & Target hypothesis for ILP\\
$C$ & Context for CDPIs in ILASP\\
$e^{inc}$ & Included set of an ILASP example or partial interpretation\\
$e^{exc}$ & Excluded set of an ILASP example or partial interpretation\\
$S$ & Stream of task executions, as sequence of kinematic features $\pazocal{K}$ and visual features $\pazocal{G}$.\\
$\Phi$ & Action identification algorithm\\
$\pazocal{L}$ & Action labels (for classification)\\
$\pazocal{F}$ & Semantic environmental features in ASP syntax\\
$\pazocal{X} (\pazocal{X}_a)$ & Set of extended action preconditions, i.e., including both preconditions and constraints (for action $a$)\\
$\pazocal{I}$ ($\pazocal{I}_F$) & Set of initiating conditions for effects of actions (for a specific environmental feature $F \in \pazocal{F}$)\\
$\pazocal{E}$ ($\pazocal{E}_F$) & Set of terminating conditions for effects of actions (for a specific environmental feature $F \in \pazocal{F}$)\\
$\rho_{ij}$ & Confidence level for timed segment $i$ to be associated with action label $l_j$\\
$\pazocal{S}$ & Set of probabilistic timed segments (after applying $\Phi$ to $S$)
\end{tabularx}

\subsection{The peg transfer task}
\begin{figure}
    \centering
    \includegraphics[scale=0.3]{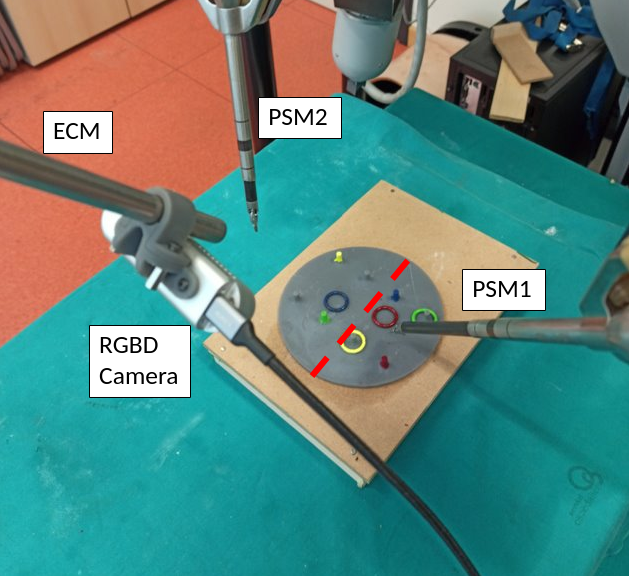}
    \caption{The setup for the peg transfer task with dVRK.}
    \label{fig:setup}
\end{figure}
Figure \ref{fig:setup} shows the setup for the illustrative surgical training task of peg transfer. The objective is to place colored rings on pegs of the corresponding color using the two patient-side manipulators (PSM1 and PSM2) of the dVRK. Each PSM can grasp any reachable ring and place it on any reachable peg; reachability is determined by the relative position of rings and pegs with respect to the center of the base. Reachability regions are identified by the red dashed line in Figure \ref{fig:setup}.
For instance, PSM1 can reach the red ring and the blue peg, while PSM2 can reach the blue ring and the green peg. 
In case an arm can reach a ring but not the same-colored peg, it must move to the center of the base and transfer the ring to the other arm to complete the task.
Pegs can be occupied by other rings and must be freed before placing the desired ring on it. Also, rings may be on pegs or on the base in the initial state, i.e., some rings may need to be extracted before being moved.

\subsection{ASP planning task description}\label{sec:asp}
We describe the peg transfer task in an established format for Answer Set Programming (ASP) by \cite{asp_syntax}. 

A (planning) domain's description in ASP comprises a \emph{sorted signature} and \emph{axioms}. 
The sorted signature is the alphabet of the domain, defining \emph{constants} (either Booleans, strings or integers), \emph{variables} and \emph{atoms}, i.e., predicates of variables.
In the peg transfer domain, variables represent arms \stt{A} (with constant value either \stt{psm1} or \stt{psm2}), objects \stt{O} (either \stt{ring} or \stt{peg}) and colors (either \stt{red, green, blue, yellow} or \stt{grey}).
Atoms represent concepts as the relative locations between objects and / or arms (e.g., \stt{at(A,ring,C)}, meaning that arm \stt{A} is at the location of ring with color \stt{C}, or \stt{placed(ring,C1,peg,C2)}, meaning that ring with color \stt{C1} is placed on peg with color \stt{C2}) and the status of grippers of arms (e.g., \stt{closed\_gripper(A)}.
In our previous work \citep{meli2021inductive} we considered more task-specific environmental concepts, e.g., \stt{in\_hand(A,R,C)} to represent that a ring is held by an arm. However, these can be derived from other atoms (for instance, \stt{in\_hand(A,R,C)} holds when both \stt{at(A,R,C)} and \stt{closed\_gripper(A)} do), so we omit them in this paper, in order to allow dependency only on basic kinematic information from the robot and preserve generalization to different tasks.
Atoms also represent actions of the task, i.e., \stt{move(A,O,C)} (moving to a colored ring or peg), \stt{move(A,center,C)} (moving to center for transfer of ring \stt{C}), \stt{grasp(A,ring,C)}, \stt{release(A)} and \stt{extract(A,ring,C)} for gripper actions and extraction, respectively.
We further annotate atoms with a temporal variable \stt{t} representing the discrete time step of the task flow (e.g., \stt{move(A,O,C,t)}).
Whenever all variables in an atom are assigned with constant values, we say that the atom is \emph{ground}.

Given the sorted signature, axioms define preconditions and effects of actions and constraints.
Preconditions for the peg transfer task (in the form \stt{Action : Precondition}) are:
\begin{subequations}
\label{eq:pre}
\begin{align}
&\stt{0 \{move(A, O, C, t) : reachable(A, O, C, t);} \\
&\stt{move(A, center, C, t) : at(A, R, C, t), closed\_gripper(A, t);}\\
&\stt{extract(A, ring, C, t) : at(A, ring, C, t), closed\_gripper(A, t);}\\
&\stt{grasp(A, ring, C, t) : at(A, ring, C, t);}\\
&\stt{release(A, t) : at(A, ring, C, t), closed\_gripper(A, t)\} 1.}
\end{align}
\end{subequations}
where the \emph{aggregate} construct \stt{0 \{..\} 1} indicates that only one action per time step can be executed, given that the corresponding precondition holds.

Effects of actions are expressed according to the generic paradigm of event calculus by \cite{kowalski1989logic}, i.e., a generic environmental atom \stt{F} is subject to the law of inertia: 
\begin{align*}
    &\stt{holds(F, t) :- initiated(F, t).} \\ \nonumber
    &\stt{holds(F, t) :- holds(F, t-1), not terminated(F, t).}
\end{align*}
where \stt{:-} equals to $\leftarrow$ (logical implication) and \stt{not} is the default negation (if a Boolean $b$ is not known to be true, i.e., unobserved, then \stt{not} $b$ is true).
The inertia law can be interpreted as follows: an environmental atom starts holding as some \stt{initiated(F)} condition holds, and ends holding at the occurrency of some \stt{terminated(F)} condition. As an example, consider the peg transfer task. Atom \stt{at(A,ring,C)} starts holding when \stt{move(A,ring,C)} has been executed, and terminates only when arm \stt{A} moves away. The inertia law is valid in every robotic scenario involving interaction with the environment.
Effects of actions in the peg transfer task are then equivalently described by the following initiating and terminating conditions:
\begin{subequations}
\begin{align}
\label{eqn:initiated_original}
    &\stt{initiated(closed\_gripper(A), t) :- grasp(A, ring, C, t-1).} \\\nonumber
    &\stt{initiated(placed(ring, C1, peg, C2), t) :- at(A, ring, C1, t-1),}\\\
    &~~~~\stt{ closed\_gripper(A, t-1), at(A, peg, C2, t-1), release(A, t-1).} \\
    &\stt{initiated(at(A, ring, C), t) :- move(A, ring, C, t-1).}\\
    &\stt{initiated(at(A, peg, C), t) :- move(A, peg, C, t-1).}\\
    &\stt{initiated(at(A, center), t) :- move(A, center, C, t-1).}\\
    &\stt{terminated(closed\_gripper(A), t) :- release(A, t-1).} \\\nonumber
    &\stt{terminated(placed(ring, C1, peg, C2), t) :- extract(A, ring, C1, t-1),}\\\label{eq:color1}
    &~~~~\stt{color(C2).} \\
    \label{eq:term_at_orig_1}&\stt{terminated(at(A, ring, C1), t) :- move(A, ring, C2, t-1), color(C1).}\\
    &\stt{terminated(at(A, ring, C1), t) :- move(A, peg, C2, t-1), color(C1).}\\
    \label{eq:term_at_orig_2}&\stt{terminated(at(A, ring, C1), t) :- move(A, center, C2, t-1), color(C1).}\\
    &\stt{terminated(at(A, peg, C1), t) :- move(A, peg, C2, t-1), color(C1).}\\
    &\stt{terminated(at(A, peg, C1), t) :- move(A, ring, C2, t-1), color(C1).}\\\label{eq:colorn}
    &\stt{terminated(at(A, peg, C1), t) :- move(A, center, C2, t-1), color(C1).}\\
    &\stt{terminated(at(A, center), t) :- move(A, O, C, t-1).}
\end{align}
\end{subequations}
The specification of \stt{color(C2)} in axioms \eqref{eq:color1}-\eqref{eq:colorn} is required in ASP to guarantee safety of rules, i.e., all variables on the left-hand side (\emph{head of rule}) of logical implication must appear without negation on the right-hand side (\emph{body of rule}).

Finally, constraints of the task are:
\begin{subequations}
\begin{align}
    \nonumber
    &\stt{:- move(A, center, C1, t), closed\_gripper(A, t), at(A, ring, C2, t),}\\\label{eq:const_movecenter} 
    &~~~~\stt{ placed(ring, C2, peg, C3, t).}\\
    \nonumber
    &\stt{:- move(A, peg, C1, t), closed\_gripper(A, t), at(A, ring, C2, t),}\\\label{eq:const_movepeg1}
    &~~~~\stt{ placed(ring, C2, peg, C3, t).}\\\label{eq:const_movering}
    &\stt{:- move(A, ring, C, t), closed\_gripper(A, t).} \\
    &\stt{:- move(A, peg, C1, t), placed(ring, C2, peg, C1, t).}\label{eq:const_movepeg2}
\end{align}
\end{subequations}
The lack of head atoms is equivalent to $\bot \leftarrow \ldots$, i.e., body atoms cannot concurrently be ground, otherwise \emph{falsum} $\bot$ would be implied.
The semantics of above constraints is then, respectively: arm \stt{A1} cannot move to the center for transfer if it is holding a ring which is still placed on a peg (extraction is required); arm \stt{A1} cannot move to a peg if it is holding a ring which is still placed on a peg; an arm with closed gripper cannot move to a ring; an arm cannot move to an occupied peg.

Given the task description, an ASP solver computes (possibly multiple) \emph{answer sets}, i.e., the minimal sets of ground atoms satisfying the specifications.
Each answer set contains actions to be executed at each time step, and the corresponding environmental context.

\subsection{Inductive learning of ASP theories}
\label{sec:ilp}
We now provide relevant definitions to introduce the problem of inductive learning of ASP theories in our considered scenario. Additional details can be found in \cite{LRB18}. 

A generic ILP problem $\pazocal{T}$ for ASP theories is defined as the tuple $\pazocal{T} = \langle B, S_M, E \rangle$, where $B$ is the \emph{background knowledge}, i.e. a set of axioms and atoms in ASP syntax; $S_M$ is the \emph{search space}, i.e. the set of candidate ASP axioms that can be learned; and $E$ is a set of \emph{examples}. The goal of $\pazocal{T}$ is to find a subset $H \subseteq S_M$ such that $H \cup B \models E$.
In ILASP \citep{law2018inductive} tool considered in this paper, examples are \emph{partial interpretations} defined as follows.
\begin{defn}[Partial interpretation]
\label{def:partial_int}
Let $P$ be an ASP program. Any set of ground atoms that can be generated from axioms in $P$ is an \emph{interpretation} of $P$. Given an interpretation $I$ of $P$, we say that a pair of subsets of grounded atoms $e = \langle e^{inc}, e^{exc} \rangle$ (respectively the included and excluded set) is a \emph{partial interpretation (extended by interpretation $I$)} if $e^{inc} \subseteq I$ and $e^{exc} \cap I = \emptyset$.
\end{defn}
\noindent
In particular, we consider \emph{context-dependent partial interpretations}, as introduced by \cite{LRB16}.
\begin{defn}[Context-dependent partial interpretation (CDPI)]
\label{def:CDPI}
A CDPI of an ASP program P with an interpretation I is a tuple $e_c = \langle e, C \rangle$, where $e$ is a partial interpretation and $C$ is an ASP program called \emph{context}. $I$ is said to extend $e_c$ if $e^{inc} \cup C \subseteq I$ and $(e^{exc} \cup C) \cap I = \emptyset$.
\end{defn}
To better understand CDPIs, consider the task of learning preconditions of action \stt{release(A,t)} in the peg transfer domain. 
Hence, $S_M$ includes axioms in the form \stt{release(A,t) :- F}$_1$\stt{, \ldots F}$_n$, being \stt{F}$_i$ a generic environmental atom of the task. 
In ILASP syntax, a CDPI may be expressed as
\begin{align}\label{eq:example_cdpi}
    &\stt{\#pos(id, \{release(psm1,1)\}, \{release(psm2,1)\}, \{at(psm1,peg,red,1), }\\\nonumber
    &~~~~\stt{closed\_gripper(psm1,1), closed\_gripper(psm2,1), \ldots\}).}
\end{align}
\noindent
where \stt{id} denotes the unique identifier of the example, $e^{inc} = \{\stt{release(psm1,1)}\}$ means that the gripper of PSM1 was open at time step 1, $e^{exc} = \{\stt{release(psm2,1)}\}$ means that the gripper of PSM2 was not open at time step 1, and $C = \{\stt{at(psm1,ring,red,1)}, \ldots\}$ is the set of ground environmental atoms at time step 1. In other words, $e^{inc}$ and $e^{exc}$ include the ground atoms corresponding to the head of the axioms to be learned, while $C$ includes all other ground atoms. 

We now specialize $\pazocal{T}$ to CDPIs as follows:
\noindent
\begin{defn}[Learning ASP theories from CDPIs]
\label{def:ILASP_CDPI}
An ILASP learning task from CDPIs is a tuple $\pazocal{T} = \langle B, S_M, E \rangle$, where $E$ is a set of CDPIs. We say that $H \subseteq S_M$ is a solution to $\pazocal{T}$ if the following holds:
\begin{equation*}
     \forall e \in E \ \exists AS \ s.t. \ B \cup H \cup C \models AS : \ e \textrm{ is extended by } AS 
\end{equation*}
\noindent
More specifically, ILASP computes the \emph{shortest} hypothesis, i.e., the one with minimal number of atoms.
\end{defn}
\noindent
For instance, from example \eqref{eq:example_cdpi}, ILASP can learn
\begin{equation}\label{eq:learn_example_cdpi}
    \stt{release(A, t) :- at(A, peg, C, t).}
\end{equation}
\noindent
where the excluded set of example \eqref{eq:example_cdpi} allows to discard the possible axiom \stt{release(A, t) :- closed\_gripper(A, t)}.

In this paper, we consider an extension of the task in Definition \ref{def:ILASP_CDPI}, i.e., learning ASP theories from \emph{noisy} CDPIs as proposed by \cite{law2018inductive}:
\noindent
\begin{defn}[Learning ASP theories from noisy CDPIs]
\label{def:ILASP_CDPI_noise}
An ILASP learning task from noisy CDPIs is a tuple $\pazocal{T} = \langle B, S_M, E \rangle$, where $E$ is a set of $N$ noisy CDPIs, i.e., CDPIs annotated with an integer weight $w_i, \forall i = 1 \ldots N$. We say that $H \subseteq S_M$ is a solution to $\pazocal{T}$ if the following holds:
\begin{equation}\label{eq:hyp}
     \forall e \in E \ \exists AS \ s.t. \ B \cup H \cup C \models AS : \ e \textrm{ is extended by } AS 
\end{equation}
More specifically, ILASP computes the hypothesis which minimizes the cost $\Gamma = \lambda + \eta$, where $\lambda$ is the number of atoms in the axioms of the hypothesis and $\eta = \sum_{i=1}^{K} w^u_i$, being $\{w^u_i\}_{i=1}^{K}$ the set of weights for $K\leq N$ uncovered examples, i.e., CDPIs not satisfying axiom \eqref{eq:hyp}.
\end{defn}
\noindent
In other words, Definition \ref{def:ILASP_CDPI_noise} introduces the possibility to ``ignore" some examples, if their weight is low. This is particularly useful to discard highly uncertain examples, e.g., resulting from poor-quality identification.
Consider, for instance, the following noisy CDPIs:
\begin{subequations}    
\begin{align}
    \nonumber&\stt{\#pos(id@}1\stt{, \{release(psm1,1)\}, \{release(psm2,1)\}, \{at(psm1,peg,red,1), }\\
    \label{eq:example_noisy1}&~~~~\stt{closed\_gripper(psm1,1), closed\_gripper(psm2,1), \ldots\}).}\\
    \nonumber&\stt{\#pos(id@}2\stt{, \{release(psm1,3)\}, \{release(psm2,3)\}, \{at(psm1,peg,red,3), }\\
    \label{eq:example_noisy2}&~~~~\stt{at(psm2,peg,blue,3), closed\_gripper(psm1,3), \ldots\}).}
\end{align}
\end{subequations}
\noindent
The first one would lead to learn
\begin{equation}\label{eq:example_goodrulenoisy}
    \stt{release(A, t) :- at(A, peg, C, t).}
\end{equation}
\noindent
while from the second one
\begin{equation}\label{eq:example_badrulenoisy}
    \stt{release(A, t) :- closed\_gripper(A, t).}
\end{equation}
\noindent
However, axiom \eqref{eq:example_badrulenoisy} has $\Gamma = 2 + 1 = 3$ (CDPI \eqref{eq:example_noisy1} is not covered), while axiom \eqref{eq:example_goodrulenoisy} has $\Gamma = 2 + 2 = 4$ (CDPI \eqref{eq:example_noisy2} is not covered). For this reason, ILASP selects \eqref{eq:example_goodrulenoisy} as the best hypothesis.

Finally, we consider another extension to Definition \ref{def:ILASP_CDPI}, including \emph{negative examples} as in \cite{SI09}.
\noindent
\begin{defn}[Learning ASP theories from positive and negative CDPIs]
\label{def:ILASP_CDPI_neg}
Consider An ILASP learning task $\pazocal{T} = \langle B, S_M, E \rangle$ as in Definition \ref{def:ILASP_CDPI}, where $E = \langle E^+, E^- \rangle$ is a set of CDPIs split into two sub-sets $E^+, E^-$, with the latter being negative and the former positive examples. We say that $H \subseteq S_M$ is a solution to $\pazocal{T}$ if the following holds:
\begin{align*}
     &\forall e \in E^+ \ \exists AS \ s.t. \ B \cup H \cup C \models AS : \ e \textrm{ is extended by } AS\\ 
     &\forall e \in E^- \ \nexists AS \ s.t. \ B \cup H \cup C \models AS : \ e \textrm{ is extended by } AS 
\end{align*}
\end{defn}
\noindent
In other words, while standard positive examples describe \emph{allowed (but not mandatory) ground atoms to be modeled by the learned ASP theory}, negative examples are more strong and describe \emph{forbidden ground atoms}, hence typically leading to the learning of task constraints. 
For instance, from example \eqref{eq:example_cdpi} ILASP may learn either axiom \eqref{eq:learn_example_cdpi} or:
\begin{equation}\label{eq:learn_aggregate}
    \stt{0 \{ release(A, t) : closed\_gripper(A, t)\} 1.}
\end{equation}
\noindent
since the aggregate construct implies that \stt{release(A, t)} \emph{may or may not be grounded}, thus $e^{inc} = \{\stt{release(psm1, 1)}\}, e^{exc}=\{\stt{release(psm2, 1)}\}$ are still allowed by Definition \ref{def:ILASP_CDPI}.
However, if we add to example \eqref{eq:example_cdpi} the negative example:
\begin{align*}
    &\stt{\#neg(id, \{release(psm2,1)\}, \{)\}, \{at(psm1,peg,red,1), }\\
    &~~~~\stt{closed\_gripper(psm1,1), closed\_gripper(psm2,1), \ldots\}).}
\end{align*}
\noindent
axiom \eqref{eq:learn_aggregate} is not allowed from Definition \ref{def:ILASP_CDPI_neg}, since it could still model $e^{inc} = \{\stt{release(psm2, 1)}\}$ which is now strictly forbidden.

\noindent

\section{Methodology}\label{sec:met}
In this section, we present our methodology for unsupervised learning and online supervised refinement of ASP task knowledge. 

\subsection{Unsupervised learning of ASP task knowledge}
The unsupervised learning strategy is sketched in Algorithm \ref{alg:learning}.
More specifically, in Section \ref{sec:dataset_gen} we describe how to generate the action-context tuples, starting from a dataset of synchronized visual and kinematic executions of the task. Then, in Section \ref{sec:learn_actions} and \ref{sec:learn_effects}, we show how to build examples from action-context tuples, in order to learn ASP axioms for constrained preconditions, and how to generate axioms for effects of actions in the event calculus, respectively.
While explaining different components of our strategy, we refer to specific lines in Algorithm \ref{alg:learning}.
\begin{algorithm}[t]
    \caption{Unsupervised learning of ASP task knowledge}\label{alg:learning}
    \begin{algorithmic}[1]
        \State \textbf{Input}: Synchronized video-kinematic stream of task executions $S$; action identification algorithm $\Phi$; set of task action labels $\pazocal{L}$; set of semantic environmental features $\pazocal{F}$
        \State \textbf{Output}: Set $\pazocal{X}$ of extended action preconditions; sets $\pazocal{I}, \pazocal{E}$ of initiating and terminating conditions, respectively, $\forall \stt{F}\in \pazocal{F}$
        \State \textbf{Initialize}: $\pazocal{X} = \pazocal{I}_\stt{F} = \pazocal{I} = \pazocal{E} = \pazocal{E}_\stt{F} = \emptyset$.
        \State $\pazocal{S} = \Phi(S)$ \% \textit{Sequence of probabilistic timed segments}
        \State $N = |\pazocal{S}|$
        \State $\pazocal{P} = \{\langle l_j, C_i, \rho_{ij} \rangle\}$ = \texttt{GENERATE\_ACT}($\pazocal{S}$) \% \textit{Generate action-context tuples}
        \For{$j \in \{1, \ldots, |\pazocal{L}|\}$}
            \For{$i \in \{1, \ldots, N\}$}
                \If{$\rho_{ij} < \frac{1}{N} \sum_{k=1}^N \rho_{kj}$}
                    \State $\pazocal{P}$.\texttt{REMOVE}($\langle l_j, C_i, \rho_{ij} \rangle$) \% \textit{Filter too noisy examples}
                \Else
                    \ForAll{\stt{F}$\in \pazocal{F}$ s.t. $\stt{f} \in G(\stt{F}) \land \stt{f} \in C_i$} \% \textit{Consider ground environmental features in the context}
                        \If{\stt{f}$\notin C_{i-1}$}
                            \State $\pazocal{I}_\stt{F}$.\texttt{APPEND}($\langle \stt{F}, l_j, \rho_{ij} \rangle$) \% \textit{Probabilistic initiating condition}
                        \EndIf
                        \If{\stt{f}$\notin C_{i+1}$}
                            \State $\pazocal{E}_\stt{F}$.\texttt{APPEND}($\langle \stt{F}, l_j, \rho_{ij} \rangle$) \% \textit{Probabilistic terminating condition}
                        \EndIf
                    \EndFor
                \EndIf
            \EndFor
            \State $\pazocal{P}_j = \{p = \langle l_j, C_i, \rho_{ij} \rangle \in \pazocal{P}\}$ \% \textit{Split action-context tuples for each action label}
            \State $\pazocal{C}$ = \texttt{GENERATE\_CDPI}($\pazocal{P}_j$) \% \textit{Generate CDPIs for each action label}
            \State $\pazocal{X}_j$ = \stt{ILASP}($\pazocal{C}$) \% \textit{Run ILASP to learn extended action preconditions for $j$-th label}
            \State $\pazocal{X}$.\stt{APPEND}($\pazocal{X}_j$)
        \EndFor
        \For{\stt{F}$\in \pazocal{F}$}
            \State $\langle \stt{F}, l^i_j, \rho_{ij} \rangle = \underset{\bar{\rho}_{j}}{\argmax}\  \pazocal{I}_{\stt{F}}$ \% \textit{Find most probable initiating condition}
            \State $\langle \stt{F}, l^e_j, \rho_{ij} \rangle = \underset{\bar{\rho}_{j}}{\argmax}\  \pazocal{E}_{\stt{F}}$ \% \textit{Find most probable terminating condition}
            \State $\pazocal{I}$.\texttt{APPEND}(\stt{initiated(F, t) :- $l_j^i$})
            \State $\pazocal{E}$.\texttt{APPEND}(\stt{terminated(F, t) :- $l_j^e$})
        \EndFor
        \Return $\pazocal{X}, \pazocal{I}, \pazocal{E}$
    \end{algorithmic}
\end{algorithm}

\subsubsection{Action-context tuples generation}\label{sec:dataset_gen}
The input to our learning pipeline is a dataset $S$ of \emph{execution traces}, i.e. synchronized kinematic and video streams of an instance of the task. Recording assumes that the robot and the vision system share a common reference frame, e.g., using the calibration procedure by \cite{roberti2020improving}.

The kinematic stream includes the following features $\pazocal{K}$: i) Cartesian coordinates of each arm $A$ of the robotic system $\vect{p}_A = \{x_{pos,A}, y_{pos,A}, z_{pos,A}\}$; ii) quaternion coordinates of each arm $A$ of the robotic system $\vect{q}_A = \{x_{or,A}, y_{or,A}, z_{or,A}, w_{or,A}\}$; iii) opening angles $j_A$ of the grippers. Other works by \cite{van2019weakly, despinoy2015unsupervised} consider additional kinematic features, e.g., Cartesian and rotational velocities; however, following our previous work \citep{meli2021unsupervised}, we decide not to consider them without loss of generality.

The video stream is acquired from a RGBD camera. The following geometric features $\pazocal{G}$ are extracted using standard color segmentation and shape recognition with random sample consensus \citep{ransac}: i) center position $\vect{p}_{rc} = \{x_{rc}, y_{rc}, z_{rc}\}$ of each ring with color $c$; ii) tip position $\vect{p}_{pc} = \{x_{pc}, y_{pc}, z_{pc}\}$ of each peg with color $c$; iii) position $\vect{p}_{b} = \{x_{b}, y_{b}, z_{b}\}$ of meeting point at the center of the base for transfer; iv) ring radius $rr$.

Kinematic and visual geometric quantities can be automatically mapped to environmental features $\pazocal{F}$ at each time step with the following relations, as proposed by \cite{IROS2020}: 
\begin{align}\label{eq:ground_fluents}
    \stt{at(A,ring,C)} &\leftarrow ||\vect{p}_A - \vect{p}_{rc}||_2 < rr\\\nonumber
    \stt{at(A,peg,C)} &\leftarrow ||\vect{p}_A - \vect{p}_{pc}||_2 < rr \ \land \ z_{pc} < z_{pos,A}\\\nonumber
    \stt{placed(ring,C1,peg,C2)} &\leftarrow ||\vect{p}_{rc1} - \vect{p}_{pc2}||_2 < rr \ \land  \ z_{rc1} < z_{pc2}\\\nonumber
    \stt{reachable(A,O,C)} &\leftarrow argmin_{A}|y_{oc} - y_{pos,A}| = A\footnote{$y_{oc}$ denotes the $y$-coordinate of entity \stt{O}$\in$\stt{\{ring, peg\}}.}\\\nonumber
    \stt{closed\_gripper(A)} &\leftarrow j_A < \frac{\pi}{8}\footnote{$\frac{\pi}{8}$ is chosen empirically to identify that the gripper is closed, since the angle is not exactly $0$ from the encoders of dVRK.}\\\nonumber
    \stt{at(A,center)} &\leftarrow ||\{x_{pos,A}, y_{pos,A}\} - \{x_{b}, y_{b}\}||_2 < rr
\end{align}
\footnotetext[4]{$y_{oc}$ denotes the $y$-coordinate of entity \stt{O}$\in$\stt{\{ring, peg\}}.}
\footnotetext[5]{$\frac{\pi}{8}$ is chosen empirically to identify that the gripper is closed, since the angle is not exactly $0$ from the encoders of dVRK.}
\noindent
The set $\pazocal{F}$ in this paper includes only commonsense information about the relative position of objects and robotic arms and the status of the gripper, hence discarding higher-level concepts as \emph{the arm is holding a ring} defined by \cite{meli2021unsupervised}. In this way, we reduce the amount of sophisticated prior knowledge required for action identification and task knowledge induction.
In fact, the definition of semantic features may be retrieved or straightforwardly derived from available domain ontologies, e.g., by \cite{gibaud2018} for surgery and \cite{Schlenoff2012} for robotics and automation (for instance, relative positions are connected to the concept of distance, retrieved from the already available \stt{PositionPoint} class in \cite{Schlenoff2012}), thus requiring minimal effort from the human designer.

We can now generate the action-context tuples needed for our learning pipeline, using any algorithm $\Phi$ for action identification from video-kinematic input.
Such algorithms split the visual-kinematic stream of execution into $N$ timed segments $\langle \pazocal{K}_i, \pazocal{G}_i, t_i \rangle$ (\underline{Line 4}), where $\pazocal{K}_i$ and $\pazocal{G}_i$ are respectively the kinematic and geometric features for the specific segment starting at $t_i$. Then, they cluster similar segments according to a distance metric $d_{ij} : \pazocal{K}_i \times \pazocal{G}_i \times \pazocal{K}_j \times \pazocal{G}_j \rightarrow \mathbb{R}^+$.

We then convert each $i$-th timed segment to $|\pazocal{L}|$ \emph{action-context tuples} (set $\pazocal{P}$) $\langle l_j, C_i, \rho_{ij} \rangle,\\ \forall j = 1, \ldots, |\pazocal{L}|$ (\underline{Line 6}), where $C_i$ is the \emph{context} set $G(\pazocal{F})$ of ground environmental features computed from $\pazocal{K}_i$ and $\pazocal{G}_i$ via relations \eqref{eq:ground_fluents}; $l_j$ is the unique identifier of an abstract (non-ground) action of the task in the set $\pazocal{L}$; $\rho_{ij} \in [0, 1]$ is the confidence level of the tuple, defined as:
\begin{equation*}
    \rho_{ij} = 1 - \frac{d_{ij}}{\sum\limits_{k=1}^{|\pazocal{L}|} d_{ik}}
\end{equation*}
\noindent
In other words, the action-context tuple states that under the context $C_i$, an action of type $l_j$ has been executed with probability $\rho_{ij}$.

In this paper, we assume that the number of actions in $\pazocal{L}$, i.e., $|\pazocal{L}|$, is known in advance.
It is reasonable, since typically the set of allowed actions of a robotic task (hence their corresponding motion primitives) is assumed as known \citep{Kim20211817, goel2019learning}, e.g., provided directly in the annotations of the benchmark JIGSAWS dataset for surgical gestures analysis \citep{gao2014jhu}. On the contrary, semantic information as the target of an action (e.g., \stt{move(A, ring, C)} has a ring with color \stt{C} as a target), its effects and preconditions are not known and must be discovered from the execution traces.
As a consequence, while the actual set of abstract action for peg transfer should be $\pazocal{L}=$\{\stt{move(A,ring,C), move(A,peg,C), move(A,center,C), release(A), grasp(A,ring,C), extract(A, ring, C)}\}, we only consider a less informative set $\pazocal{L}=\{l_j(\pazocal{A})\}, \forall j = 1, \ldots, |\pazocal{L}|$. where $\pazocal{A}$ is a subset of the available robotic arms ($|\pazocal{A}|\leq 2$ in peg transfer) actually involved in the specific action execution. The set $\pazocal{A}$ can be easily retrieved from the kinematic features, e.g., evaluating the variation of $\vect{p}_A$ and $\vect{q}_A$ from the beginning to the end of each timed segment returned from the identification algorithm.

\subsubsection{Constraints as extended action preconditions}
The ILASP formulation considered for unsupervised task knowledge learning reflects Definition \ref{def:ILASP_CDPI_noise}, i.e., excluding negative examples. Though it is possible to define noisy negative examples and extend Definition \ref{def:ILASP_CDPI_neg} similarly to Definition \ref{def:ILASP_CDPI_noise}, this would not be a correct modeling approach. In fact, we learn task knowledge from examples generated by an unsupervised action identification algorithm, which only considers \emph{observed executions of specific actions}, without any information about not executed actions. Hence, similarly to our previous works on discovering procedural knowledge from uncertain execution \citep{mazzi2023learning,meli2024learning}, our task mainly matches the definition of positive examples with an included (observed) and and excluded (unobserved) set in the CDPIs, rather than introducing strictly forbidding negative examples. 

At the same time, only negative examples allow to learn hard constraints typical of real robotic tasks. For this reason, we embed constraints as \emph{extended preconditions} of actions. In fact, constraints \eqref{eq:const_movecenter}-\eqref{eq:const_movepeg2} can be added to preconditions \eqref{eq:pre} as:
\begin{subequations}
\label{eq:ext_pre}
\begin{align}
\label{eq:ext_pre_movering}
&\stt{0 \{move(A,ring,C,t) : reachable(A,ring,C,t),not closed\_gripper(A,t);} \\
\label{eq:ext_pre_movepeg1}&\stt{move(A,peg,C,t) : reachable(A,peg,C,t),not placed(ring,C1,peg,C,t);} \\\nonumber
&\stt{move(A,peg,C,t) : reachable(A,peg,C,t), not placed(ring,C1,peg,C2,t),}\\
\label{eq:ext_pre_movepeg2}&~~~~\stt{closed\_gripper(A,t),at(A,ring,C1,t);} \\\nonumber
&\stt{move(A,center,C,t) : at(A,ring,C,t),not placed(ring,C,peg,C1,t),}\\
\label{eq:ext_pre_movecenter}&~~~~\stt{closed\_gripper(A,t),at(A,ring,C1,t);} \\
&\stt{extract(A,ring,C,t) : at(A,ring,C,t),closed\_gripper(A,t);}\\
&\stt{grasp(A,ring,C,t) : at(A,ring,C,t);}\\
&\stt{release(A,t) : at(A,ring,C,t),closed\_gripper(A,t)\} 1.}
\end{align}
\end{subequations}
\noindent
where constraint \eqref{eq:const_movecenter} is embedded in axiom \eqref{eq:ext_pre_movecenter}, constraints \eqref{eq:const_movepeg1}-\eqref{eq:const_movepeg2} are represented in \eqref{eq:ext_pre_movepeg1}-\eqref{eq:ext_pre_movepeg2} and constraint \eqref{eq:const_movering} is added to precondition \eqref{eq:ext_pre_movering}.

\subsubsection{Learning extended action preconditions}\label{sec:learn_actions}
We now show how to translate action-context tuples into CDPIs for learning ASP task knowledge.
We assume that axioms for each action can be learned separately, i.e., it is possible to launch ILP tasks in Definition \ref{def:ILASP_CDPI_noise} in parallel for all actions.
This is reasonable, since we want to capture ASP axioms matching partial interpretations to the context, i.e., relations between actions and environmental features similar to the ones presented in Section \ref{sec:asp}. Hence, mutual dependencies between actions can be ignored, since they are implicitly represented with effects of actions on the environmental context.
Moreover, we discard $i$-th action-context tuple for a given action label \stt{l}$_j$ (\underline{Lines 9-10}) if 
\begin{equation}\label{eq:filter_noise}
    \rho_{ij} < \frac{1}{N} \sum_{k=1}^N \rho_{kj}
\end{equation}
\noindent
In this way, we reduce the input noise to ILASP.

For each $i,j$-th action-context tuple, we then build a noisy CDPI (\underline{Line 18}) in the form:
\begin{equation}\label{eq:cdpi_exc}
    \stt{\#pos(id@}\rho_{ij}\stt{, }G(l_j)\stt{, } \pazocal{H}(l_j) \setminus G(l_j)\stt{, } C_i\stt{).}
\end{equation}
\noindent
where $\pazocal{H}(l_j)$ is the Herbrand base of $l_j = l_{j}(\pazocal{A})$, i.e., the set of all possible ground atoms which can be derived from $l_j$ (i.e., the different possible assignments of arm variables in $\pazocal{A}$); $G(l_j) \subseteq \pazocal{H}(l_j)$ is the set of ground atoms actually observed at the specific time step represented by the example. 
For instance, if PSM1 moves towards the red ring, $l_j$ corresponds to \stt{move(A, ring, C) = }$l_j$\stt{(A)},  $G(l_j) = \{l_j$\stt{(psm1)}\}, $\pazocal{H}(l_j) = \{l_j$\stt{(psm1)}, $l_j$\stt{(psm2)}\}.
The set $e^{exc}$ allows to learn extended preconditions embedding constraints.

Given these CDPIs, for each $l_j$ we define an ILASP task (\underline{Line 19}) to learn extended action preconditions (set $\pazocal{X}$), with background knowledge containing the definition of $\pazocal{F}$ and ASP variables (see Section \ref{sec:asp}) and the search space consisting of axioms in the form $l_j$\stt{ :- F}$_1, \ldots$, \stt{F}$_n$, being \stt{F}$_i \in \pazocal{F}, \forall i = 1, \ldots, n$.

\subsubsection{Generating effects of actions}\label{sec:learn_effects}
The quality of axioms representing effects of actions depends on the accuracy in the identification of action labels.
We generate axioms corresponding to effects of actions as follows.
Assuming the $i,j$-th action-context tuple represents the task situation at timestep $t$, with context $C_t \equiv C_i$, we evaluate the differences between $C_{t-1}$-$C_{t}$ on one side, and $C_{t+1}$-$C_t$ on the other side\footnote{In Algorithm \ref{alg:learning}, we denote the temporal subscript as $i$ in the definition of the contexts (\underline{Lines 12-16}). In fact, from \underline{Line 5} and \underline{Line 8}, $i \in 1, \ldots, N$. On the contrary, in the main text of the paper, we use $t-1$ and $t+1$ to refer to previous and following temporal instants, since we assume some action-context tuples have already been removed by criterion \eqref{eq:filter_noise}, hence the sequence of $i$'s does not necessarily match the temporal sequence.}.
Specifically, for each ground atom $\stt{f} \in G(\stt{F})$, with \stt{F}$\in \pazocal{F}$, we verify:
\begin{itemize}
    \item if \stt{f}$\in C_t$ and \stt{f}$\notin C_{t-1}$, we define a \emph{probabilistic initiating example} $\langle \stt{F}, l_j, \rho_{ij} \rangle$, meaning that the effect of $l_j$ is to ground the environmental atom \stt{F} with probability $\rho_{ij}$ (\underline{Lines 13-14});
    \item if \stt{f}$\in C_t$ and \stt{f}$\notin C_{t+1}$, we define a \emph{probabilistic terminating example} $\langle \stt{F}, l_j, \rho_{ij} \rangle$, meaning that the effect of $l_j$ is to unground the environmental atom \stt{F} with probability $\rho_{ij}$ (\underline{Lines 15-16}).
\end{itemize}
\noindent
For each \stt{F}, we obtain a set of probabilistic initiating and terminating examples $\pazocal{I}_{\stt{F}}$, $\pazocal{E}_{\stt{F}}$, respectively.
We then rank these sets according to 
\begin{equation*}
    \bar{\rho}_{j} = \frac{1}{K} \sum\limits_{i=1}^K \rho_{ij} \quad \forall \langle \stt{F}, l_j, \rho_{ij} \rangle \in \pazocal{I}_{\stt{F}} \ (\mathrm{respectively, } \ \pazocal{E}_{\stt{F}})
\end{equation*}
\noindent
The above mean measures the accuracy of identifying a generic action $l_j$ as an initiating or terminating condition for a semantic feature \stt{F}.
We then identify $l_j^e$ and $l_j^i$, corresponding respectively to $\max_{\bar{\rho}_{j}} \pazocal{E}_{\stt{F}}$ and $\max_{\bar{\rho}_{j}} \pazocal{I}_{\stt{F}}$ (\underline{Lines 22-23}). 
We finally define the following starting and ending conditions for \stt{F} (\underline{Lines 24-25}):
\begin{align*}
    &\stt{initiated(F, t) :- } l_j^i.\\
    &\stt{terminated(F, t) :- } l_j^e.
\end{align*}
\noindent
whose collection constitutes the sets $\pazocal{I}, \pazocal{E}$ of initiating and terminating conditions, respectively, describine the effects of robotic actions on the environment.

\subsection{Online supervised refinement of ASP task knowledge}\label{sec:met-refine}
Learned preconditions $\pazocal{X}$ and initiating / terminating conditions ($\pazocal{I}$ / $\pazocal{E}$) for effects of actions may be incomplete or inexact, depending on the quality of the input video-kinematic stream $S$ and the accuracy of the chosen unsupervised action identification algorithm $\Phi$.
Hence, especially in safety-critical scenarios as robotic surgery, we must design a \emph{shared autonomy framework} where autonomous ASP reasoning (based on learned task knowledge) is interleaved with human control. At the same time, shared autonomy is useful to \emph{continuously improve} the autonomous system, exploiting human feedback as a teacher.

Two main assumptions underlie our methodology: i) ASP action labels are directly mapped to motion primitives for actual robotic execution; ii) a perception system (in our case, a RGB-D camera and kinematic sensors from the robot) is available to automatically ground semantic features $\pazocal{F}$ and perform reasoning.
Both are reasonable. In fact, once timed segments are clustered according to action labels, it is possible to efficiently learn in closed form the corresponding primitive for each action, e.g., as shown by \cite{IROS2020, ginesi2021overcoming}. 
Moreover, semantic features are simple geometric re-interpretations of kinematic and geometric features from sensors, according to relations \eqref{eq:ground_fluents}, and \cite{IROS2020, tagliabue2022deliberation, meli2021autonomous} showed that they can be efficiently computed for online planning.

We now detail the main steps of our proposed methodology.
First, preconditions in $\pazocal{X}$ are converted to a set of aggregate preconditions $\pazocal{X}_a$:
\begin{align}\label{eq:agg_trans}
    &\stt{0 \{}l_0 \stt{: F}_1, \stt{F}_2, \ldots, \stt{F}_n;\\
    \nonumber&\ \ \ \ l_1 \stt{: F}_1, \stt{F}_2, \ldots, \stt{F}_n;\\
    \nonumber&\ \ \ \ \ldots\\
    \nonumber&\ \ \ \ l_{|\pazocal{L}|} \stt{: F}_1, \stt{F}_2, \ldots, \stt{F}_n \stt{\} 1.}
\end{align}
\noindent
where \stt{F$_i \in \pazocal{F}, \forall i=1, \ldots, n$}.
In this way, we set up an ASP program combining $\pazocal{E}, \pazocal{I}, \pazocal{X}_a$ which can be used for online task planning.

At the beginning of the task, the perception system grounds semantic features, thus triggering ASP reasoning.
The generated plan is a sequence of actions $S_a = \langle \stt{a}_1, \stt{a}_2, \ldots, \stt{a}_n \rangle$, selected for execution according to a FIFO policy.
Before executing $\stt{a}_1$, which is an instance of some $l_j$, the user is asked a feedback on whether he approves it or not.
In case of approval, we generate and store a positive \emph{not noisy} CDPI for $l_j$:
\begin{equation*}
    \stt{\#pos(id}, \{\stt{a}_1\}, \pazocal{H}(l_j) \setminus \{\stt{a}_1\}, C_t\stt{).}
\end{equation*}
\noindent
being $C_t$ the set of ground environmental features at current time $t$ perceived by sensors.
Otherwise, we generate and store a negative not noisy example:
\begin{equation*}
    \stt{\#neg(id}, \{\stt{a}_1\}, \{\}, C_t\stt{).}
\end{equation*}
\noindent
We then ask the user for a feasible action $\stt{a}_u$, which is an instance of, e.g., $l_i, i\neq j$, and store a positive CDPI:
\begin{equation*}
    \stt{\#pos(id}, \{\stt{a}_u\}, \pazocal{H}(l_i) \setminus \{\stt{a}_u\}, C_t\stt{).}
\end{equation*}
\noindent
Then, we execute an action $\stt{a}^*$ (either $\stt{a}_u$ or $\stt{a}_j$), transiting to time step $t+1$.
At this point, we evaluate each ground environmental feature \stt{f} occurring in $C_t$ but not in $C_{t+1}$ (respectively, viceversa), and generate a positive CDPI for terminating (respectively, initiating) conditions of \stt{F = $G^{-1}$(f)} (i.e., the corresponding not ground semantic feature) as:
\begin{align*}
    &\stt{\#pos(id, \{terminated(f, t+1)\}, \{\}, }C_t \setminus \stt{\{f\}} \cup \{\stt{a}^*\}\stt{).}\\
    &\textrm{respectively}\\
    &\stt{\#pos(id, \{initiated(f, t+1)\}, \{\}, }C_t \setminus \stt{\{f\}} \cup \{\stt{a}^*\}\stt{).}
\end{align*}
\noindent
After an action belonging to $l_j$ has been \emph{forbidden}, and an alternative one belonging to $l_i$ has been provided by the user, we run ILASP online for both action labels, as well as the initiating and terminating conditions for all newly grounded or ungrounded semantic features. We then update corresponding axioms in $\pazocal{E}, \pazocal{I}, \pazocal{X}_a$.
In this way, we do not refine task knowledge at each step of execution, which would be too computationally demanding, but only when the autonomous system exhibits unwanted, potentially unsafe behavior (i.e., available task knowledge is not correct).

\section{Results}\label{sec:res}
We now validate our methodology in the peg transfer domain.
Specifically, we first analyze the performance of Algorithm \ref{alg:learning} on a relatively small dataset containing both visual and kinematic information about task execution.
Then, we show how to practically include learned task knowledge into an autonomous robotic framework as the one by \cite{IROS2020}, in order to realize shared autonomy with incremental refinement of ASP task knowledge.

All experiments are run on a computer with AMD Ryzen 7 3700x CPU (8 cores, 16 threads, \SI{4.4}{GHz} nominal frequency) and \SI{32}{GB} RAM.

\subsection{Unsupervised learning of ASP task knowledge}\label{sec:res-learn}
\begin{figure}
    \centering
    \begin{subfigure}{0.3\textwidth}
    \includegraphics[height=0.9\linewidth]{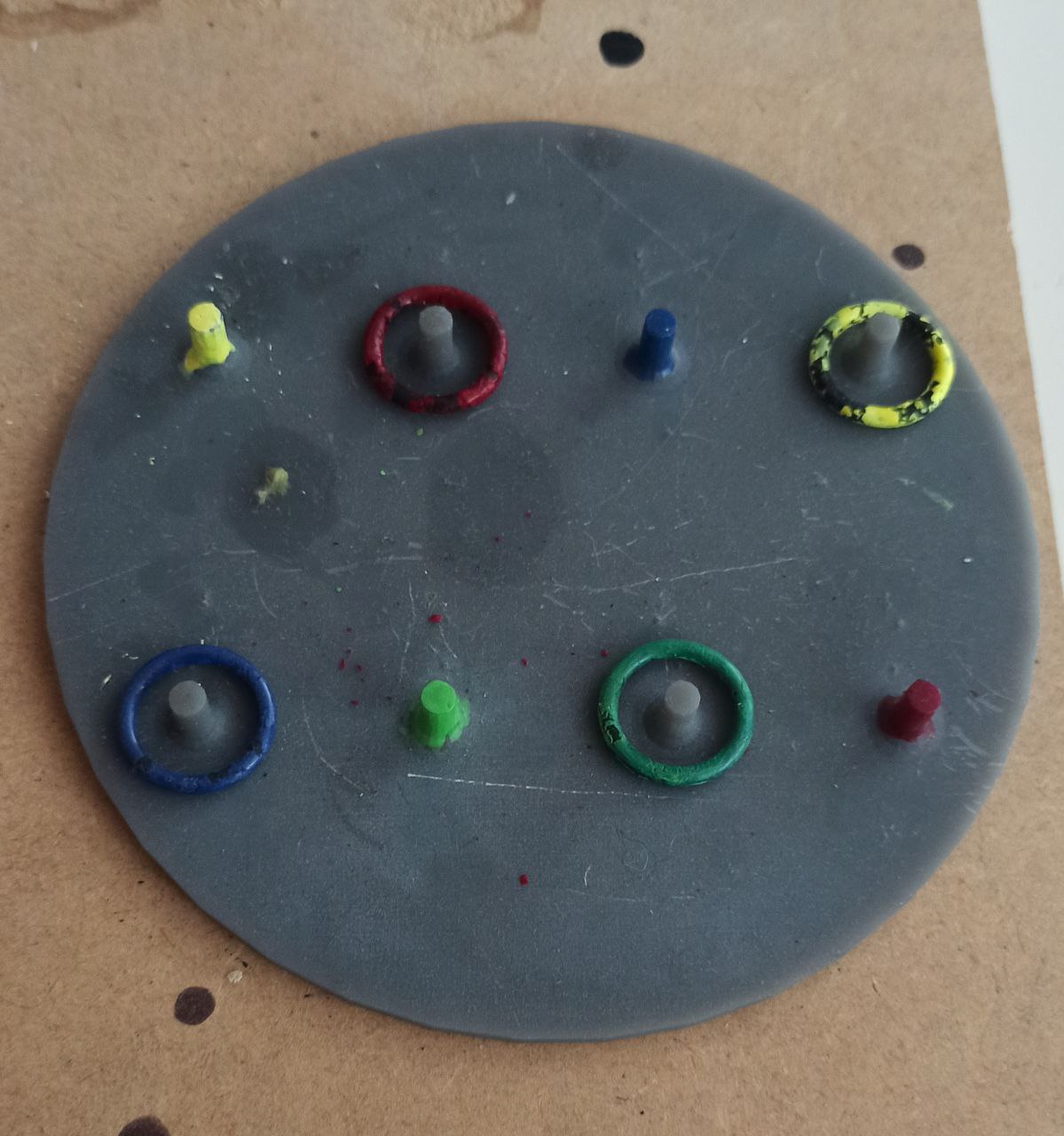}
    \caption{}
    \label{subfig:exec1}
    \end{subfigure}
    \begin{subfigure}{0.3\textwidth}
    \includegraphics[height=0.9\linewidth]{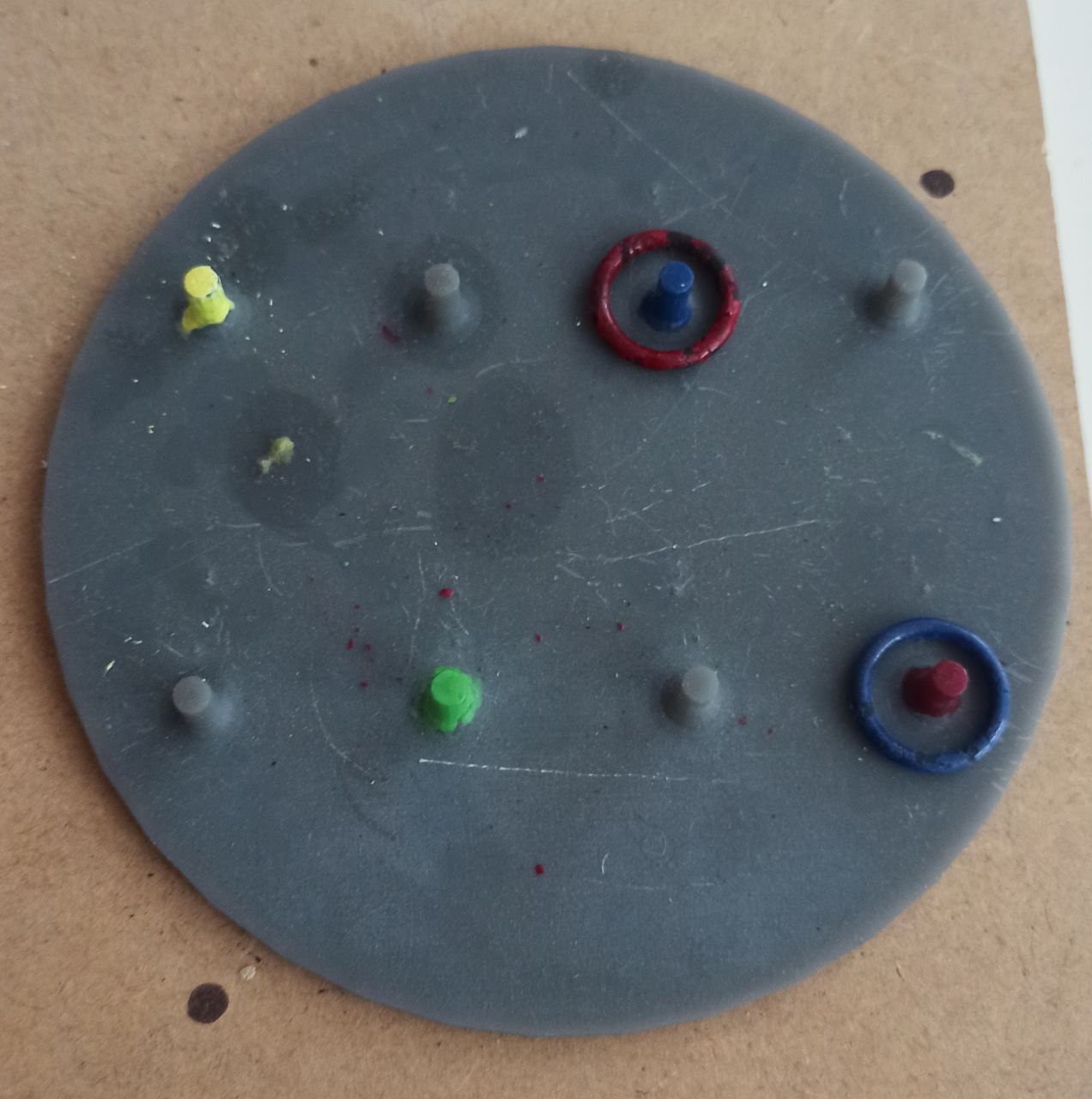}
    \caption{}
    \label{subfig:exec2}
    \end{subfigure}
    \begin{subfigure}{0.3\textwidth}
    \includegraphics[height=0.9\linewidth]{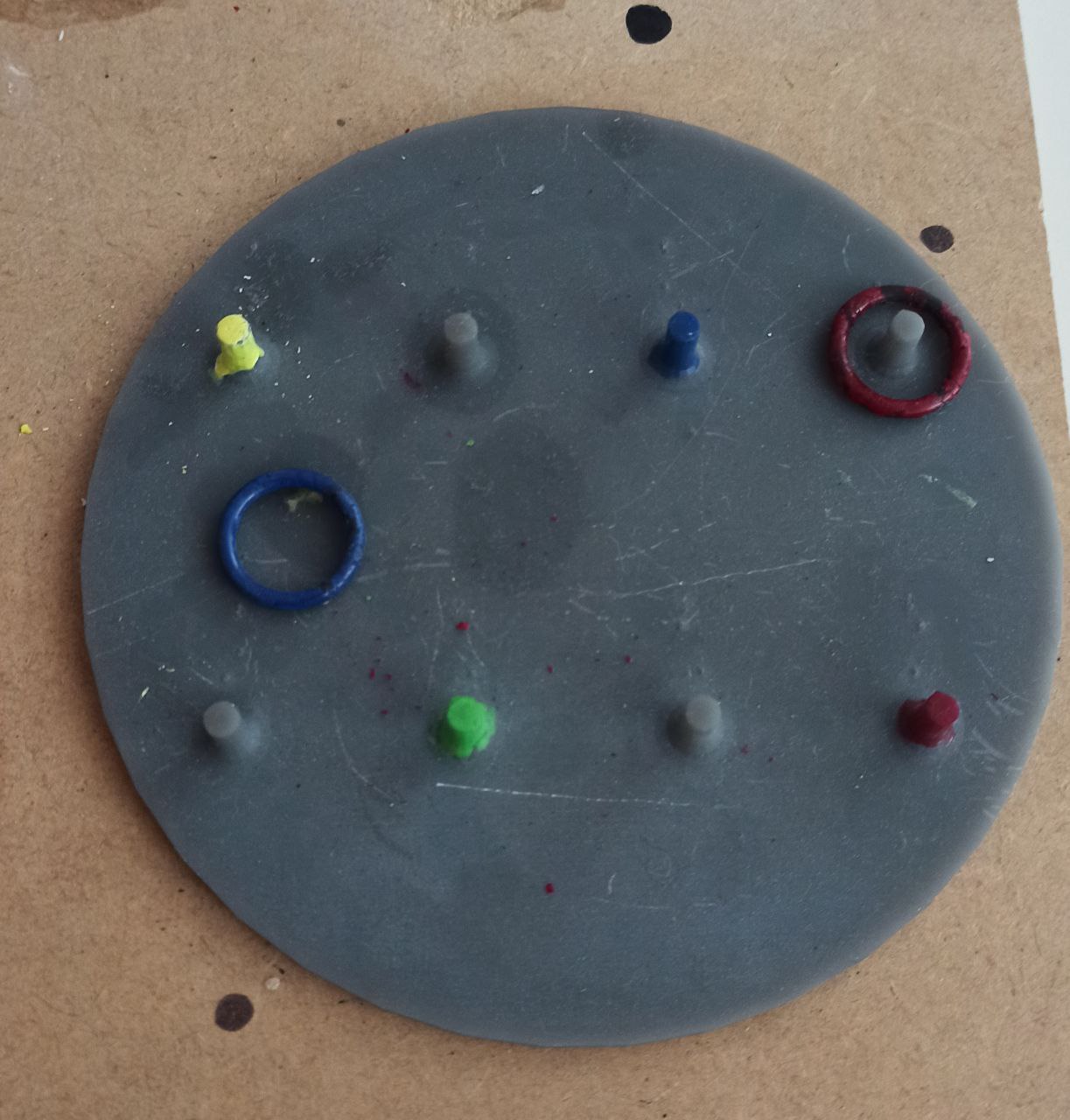}
    \caption{}
    \label{subfig:exec3}
    \end{subfigure}
    \caption{Initial environmental conditions for the task executions used for unsupervised learning.}
    \label{fig:execs}
\end{figure}

We generate an input video-kinematic stream $S$ for Algorithm \ref{alg:learning}, from the three task instances depicted in Figure \ref{fig:execs}, executed by a user in teleoperation with dVRK and using Intel Realsense RGB-D camera for visual acquisition. 
In Figure \ref{subfig:exec1}, the task starts in the nominal situation described in FLS \citep{FLS}, with all 4 colored rings placed on grey pegs, on the opposite side with respect to the corresponding pegs. Hence, PSMs must perform transfer before placing rings on pegs.
In Figure \ref{subfig:exec2}, a more challenging situation is depicted. Relevant rings and pegs (i.e., the blue and red ones) are on one side of the peg base, hence no transfer is needed. However, rings occupy pegs with mismatched color, thus one of the colored peg must be freed (by placing a ring on a grey peg temporarily), before completing the task.
Figure \ref{subfig:exec3} shows an unconventional scenario with the red ring on the same side as its peg, while the blue ring is on the opposite side out of any peg. Hence, not all rings require extraction in this task instance. Furthermore, before placing the blue ring on its peg (hence, after transfer), the ring fell, hence it had to be grasped again to complete the task.
These three task examples well represent different flows of executions of peg transfer. However, their major challenge lies in the fact that \emph{actions are not equally represented}. For instance, \stt{move(A, center, C)} occurs only 5 times (4 times in Figure \ref{subfig:exec1} and once in Figure \ref{subfig:exec3}), while \stt{release(A)} occurs 21 times (10 times in Figure \ref{subfig:exec1}, 5 times in Figure \ref{subfig:exec2} and 6 times in Figure \ref{subfig:exec3}).

We evaluate the performance of Algorithm \ref{alg:learning} in terms of standard \emph{precision, recall} and \emph{F1-score} achieved by learned axioms, given a random environmental context for the task. We assess the performance of the algorithm when using different unsupervised action identification strategies $\Phi$. 
In particular, we consider our methodology proposed in \cite{meli2021unsupervised} and the original work by \cite{despinoy2015unsupervised}, achieving lower F1-score, especially on small heterogeneous datasets of executions (as the one in Figure \ref{fig:execs}).
We refer the reader to the two references for more details. 
The essential difference between \cite{meli2021unsupervised} and \cite{despinoy2015unsupervised} is that the latter only consider the kinematic signature to cluster action segments, while the former also take into account the environmental representation given by the semantic features. Hence, \cite{meli2021unsupervised} are more robust to heterogeneous motions (a peculiarity of our dataset).
For this reason, from now on we refer to the variant with the algorithm by \cite{despinoy2015unsupervised} as $A_{kin}$, while the variant with the algorithm by \cite{meli2021unsupervised} as $A_{sem}$.
Moreover, we also consider a \emph{baseline} version of our algorithm where we perform action identification as in \cite{meli2021unsupervised} and \emph{do not consider} the excluded set in CDPIs of equation \eqref{eq:cdpi_exc}. In other words, we learn task knowledge only from observed examples of execution, neglecting counterexamples which lead to discovering constraints (extended action preconditions).
We expect the performace to degrade with the accuracy of the underlying action identification strategy.

In order to evaluate learned axioms for extended action preconditions by each algorithm version, we generate 1000 random contexts (i.e., ground semantic environmental features) for the task. For each context, we then compute feasible ground actions (only one time step ahead in the future, hence immediate actions implied by the context), performing ASP reasoning on both handcrafted task knowledge (Section \ref{sec:asp}) and learned axioms. Whenever the latter match the implications of the former, a \emph{true positive} ($TP$) is counted. Otherwise, a \emph{false positive} ($FP$) is considered. We thus compute precision, recall and F1-score as:
\begin{equation*}
    Pr = \frac{TP}{TP + FP}\hfill
    Rec = \frac{TP}{TP + FN}\hfill
    F1 = 2\frac{Pr \cdot Rec}{Pr + Rec}
\end{equation*}
\noindent
To evaluate learned effects of actions, we perform analogous experiments, but considering 1000 random action-context pairs, i.e., the same random contexts as above paired with one possible action each. 
In all tests, we also randomize the number of reachable rings $\in \{1, 2, 3, 4\}$.

For axioms related to actions, ILASP is run setting a maximum length of axioms up to 6 (i.e., 1 head atom and up to 5 body atoms), resulting in $\approx 10200$ axioms in the search space per action (learning tasks for actions are run in parallel, since they are mutually independent).
As outlined in Algorithm \ref{alg:learning}, effects of actions are not learned via ILASP.

We report the learned axioms with the three considered algorithms in Appendix \ref{sec:offline}.
Learning times in Table \ref{table:learn_time} show that the baseline algorithm is the fastest one, since only executed actions are presented to ILASP, thus the tool does not take into account extended action preconditions (constraints on actions), resulting in simpler yet more incomplete task knowledge.
Algorithm $A_{sem}$ requires less time than $A_{kin}$. In fact, the latter have a lower action identification accuracy, thus examples are more noisy and ILASP converges more slowly to a final hypothesis. The only exception is for \stt{move(A, ring, C)}, which is learned within \SI{4}{s} when using \cite{despinoy2015unsupervised}. However, in this case ILASP actually finds the empty hypothesis. 
For this reason, in Appendix \ref{app:offline_desp} we introduce the trivial axiom:
\begin{equation*}
    \stt{move(A, ring, C, t) :- arm(A), color(C).}
\end{equation*}
\noindent
meaning that moving to a ring is always possible.

Table \ref{table:results-learn} shows the quantitative results regarding the quality of learned axioms. For each metric, we report median and Inter-Quartile Range (IQR).
We highlight the best results in bold (we focus first on the highest median, then on the lowest IQR).
When using $A_{sem}$, the F1-score is the highest in most cases. 
When evaluating effects of actions (statistics about \stt{initiating} and \stt{terminating} conditions are merged), the baseline algorithm achieves the same performance as $A_{sem}$. In fact, as explained in Section \ref{sec:learn_effects}, the excluded set introduced in CDPIs in equation \eqref{eq:cdpi_exc} for learning action preconditions does not affect the way we learn axioms for effects. Thus, the main element affecting the results for effects of actions is the chosen unsupervised action identification algorithm.
The baseline algorithm achieves the worst performance at learning action preconditions for grasping, releasing and extracting. In fact, these actions involve repetitive fast motions, either opening / closing of the gripper or a simple motion normal to the peg base. On the contrary, in our non-homogeneous dataset, moving actions correspond to different shapes of motion, depending on the specific locations of rings and pegs. Thus, the kinematic signature of action segments is more heterogeneous and induces higher errors in \cite{despinoy2015unsupervised}. For this reason, the baseline algorithm still achieves better results than $A_{kin}$.

\begin{table}[]
    \centering
    \begin{tabular}{c|c|c|c}
        \toprule
        Head & $\mathbf{A_{sem}}$ & $A_{kin}$ & Baseline\\
        \midrule
         \stt{move(A, ring, C)} & 69 & 4 & 34\\
         \stt{move(A, peg, C)} & 42 & 37 & 12\\
         \stt{move(A, center, C)} & 55 & 151 & 26\\
         \stt{grasp(A, ring, C)} & 20 & 141 & 12\\
         \stt{extract(A, ring, C)} & 44 & 127 & 12\\
         \stt{release(A)} & 166 & 187 & 81\\
         \midrule
         \textbf{Average} & 66 & 108 & 30 \\
         \bottomrule
    \end{tabular}
    \caption{Learning time (in seconds) required by ILASP to discover axioms for actions from the dataset of executions depicted in Figure \ref{fig:execs}. Values are truncated to the units since they are collected from a single execution of ILASP for each action.}
    \label{table:learn_time}
\end{table}

\begin{table}[t]
\centering
    \caption{Quantitative results (median - IQR range) of different variants of Algorithm \ref{alg:learning}. We aggregate \stt{initiated} and \stt{terminated} conditions for effects of actions.}
    \label{table:results-learn}
    \begin{tabular*}{\hsize}{@{\extracolsep{\fill}}c|c|c|c|c@{}}
        \toprule
        Head & & $\mathbf{A_{sem}}$ & $A_{kin}$ & Baseline\\
        \midrule
        \multirow{3}{*}{\stt{move(A, ring, C)}} & $Pr$ & $1.00 - 0.00$ & $1.00 - 0.00$ & $0.50 - 1.00$\\  & $Rec$ & $0.50 - 0.50$ & $0.00 - 1.00$ & $1.00 - 0.00$\\ & $F1$ & $\mathbf{0.67 - 0.33}$ & $0.00 - 1.00$ & $0.67 - 0.80$\\
        \midrule
        \multirow{3}{*}{\stt{move(A, peg, C)}} & $Pr$ & $1.00 - 0.00$ & $1.00 - 0.00$ & $0.50 - 1.00$\\  & $Rec$ & $1.00 - 0.75$ & $1.00 - 0.75$ & $1.00 - 0.00$\\  & $F1$ & $0.40 - 1.00$ & $0.40 - 1.00$ & $\mathbf{0.67 - 1.00}$\\
        \midrule
        \multirow{3}{*}{\stt{move(A, center, C)}} & $Pr$ & $1.00 - 0.00$ & $1.00 - 1.00$ & $0.13 - 1.00$\\  & $Rec$ & $1.00 - 1.00$ & $1.00 - 1.00$ & $1.00 - 0.00$\\  & $F1$ & $\mathbf{1.00 - 1.00}$ & $0.00 - 0.40$ & $0.22 - 1.00$\\
        \midrule
        \multirow{3}{*}{\stt{grasp(A, ring, C)}} & $Pr$ & $1.00 - 0.00$ & $1.00 - 0.50$ & $1.00 - 1.00$\\  & $Rec$ & $1.00 - 0.00$ & $1.00 - 0.00$ & $1.00 - 0.00$\\  & $F1$ & $\mathbf{1.00 - 0.00}$ & $1.00 - 0.33$ & $1.00 - 1.00$\\
        \midrule
        \multirow{3}{*}{\stt{extract(A, ring, C)}} & $Pr$ & $1.00 - 0.00$ & $1.00 - 0.00$ & $0.00 - 1.00$\\  & $Rec$ & $1.00 - 0.00$ & $1.00 - 0.00$ & $1.00 - 0.00$\\  & $F1$ & $\mathbf{1.00 - 0.00}$ & $\mathbf{1.00 - 0.00}$ & $0.00 - 1.00$\\
        \midrule
        \multirow{3}{*}{\stt{release(A)}} & $Pr$ & $1.00 - 0.00$ & $1.00 - 0.00$ & $0.50 - 1.00$\\  & $Rec$ & $1.00 - 1.00$ & $1.00 - 0.00$ & $1.00 - 0.00$\\  & $F1$ & $1.00 - 1.00$ & $\mathbf{1.00 - 0.33}$ & $0.67 - 1.00$\\
        \midrule
        \multirow{3}{*}{\stt{closed\_gripper(A)}} & $Pr$ & $1.00 - 0.00$ & $1.00 - 0.00$ & $1.00 - 0.00$\\  & $Rec$ & $1.00 - 0.00$ & $1.00 - 0.00$ & $1.00 - 0.00$\\  & $F1$ & $\mathbf{1.00 - 0.00}$ & $\mathbf{1.00 - 0.00}$ & $\mathbf{1.00 - 0.00}$\\
        \midrule
        \multirow{3}{*}{\stt{at(A, center)}} & $Pr$ & $1.00 - 0.00$ & $1.00 - 0.00$ & $1.00 - 0.00$\\  & $Rec$ & $0.67 - 1.00$ & $0.67 - 1.00$ & $0.67 - 1.00$\\  & $F1$ & $\mathbf{0.75 - 1.00}$ & $\mathbf{0.75 - 1.00}$ & $\mathbf{0.75 - 1.00}$\\
        \midrule
        \multirow{3}{*}{\stt{at(A, ring, C)}} & $Pr$ & $1.00 - 0.00$ & $1.00 - 1.00$ & $1.00 - 0.00$\\  & $Rec$ & $1.00 - 0.88$ & $1.00 - 0.13$ & $1.00 - 0.88$\\  & $F1$ & $\mathbf{1.00 - 0.78}$ & $1.00 - 1.00$ & $\mathbf{1.00 - 0.78}$\\
        \midrule
        \multirow{3}{*}{\stt{at(A, peg, C)}} & $Pr$ & $1.00 - 0.00$ & $1.00 - 0.00$ & $1.00 - 0.00$\\  & $Rec$ & $0.75 - 0.88$ & $0.75 - 0.88$ & $0.75 - 0.88$\\  & $F1$ & $\mathbf{0.83 - 0.78}$ & $\mathbf{0.83 - 0.78}$ & $\mathbf{0.83 - 0.78}$\\
        \midrule
        \multirow{3}{*}{\stt{placed(ring, C1, peg, C2)}} & $Pr$ & $1.00 - 0.00$ & $1.00 - 1.00$ & $1.00 - 0.00$\\  & $Rec$ & $1.00 - 0.00$ & $1.00 - 0.00$ & $1.00 - 0.00$\\  & $F1$ & $\mathbf{1.00 - 0.00}$ & $0.03 - 1.00$ & $\mathbf{1.00 - 0.00}$\\
        \midrule
        \textbf{Mean F1} & & \textbf{0.88} & 0.64 & 0.73\\
        \bottomrule
    \end{tabular*}
\end{table}

\subsection{Online supervised refinement of ASP task knowledge}
\begin{figure}
    \centering
    \begin{subfigure}{0.3\textwidth}
    \includegraphics[height=0.9\linewidth]{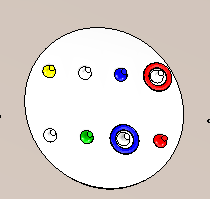}
    \caption{}
    \label{subfig:sim1}
    \end{subfigure}
    \begin{subfigure}{0.3\textwidth}
    \includegraphics[height=0.9\linewidth]{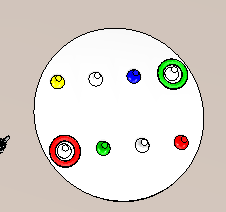}
    \caption{}
    \label{subfig:sim2}
    \end{subfigure}
    \begin{subfigure}{0.3\textwidth}
    \includegraphics[height=0.9\linewidth]{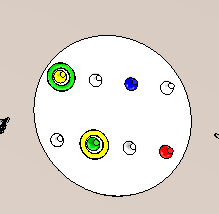}
    \caption{}
    \label{subfig:sim3}
    \end{subfigure}
    \caption{Initial environmental conditions for the task simulations used for task knowledge refinement.}
    \label{fig:sims}
\end{figure}

Learned axioms from Algorithm \ref{alg:learning} may be incomplete and unsafe for practical application in a fully autonomous robotic framework, due to inaccuracies of unsupervised action identification and Algorithm \ref{alg:learning} itself.
We now show how our shared autonomy framework (Section \ref{sec:met-refine}) can be applied to ensure safety and realize incremental refinement of learned ASP task knowledge, embedding (expert) human feedback in the reasoning and execution loop.
In particular, we investigate how precision, recall and F1-score of axioms evolve, as they are refined through expert feedback during the execution of the task (for fair comparison, we do not consider the aggregate transformation in equation \ref{eq:agg_trans}, i.e., we consider \emph{all possible ground atoms from learned axioms}).

We use a simulated model of the peg transfer scenario with dVRK and a RGB-D sensor\footnote{Calibration between the PSMs and the camera is assumed as a prior, achieved, e.g., via the methodology proposed by \cite{roberti2020improving}.}, based on the Robot Operating System\footnote{\href{https://ros.org}{https://ros.org}} and built in CoppeliaSim environment\footnote{\href{https://www.coppeliarobotics.com/}{https://www.coppeliarobotics.com/}} and shown in Figure \ref{fig:sims}.
As explained in Section \ref{sec:met-refine}, we assume that motion primitives for each action class are preliminarily learned from clusters generated from the unsupervised action identification algorithm by \cite{meli2021unsupervised}, using, e.g., dynamic movement primitives proposed by \cite{IROS2020}\footnote{An user may also specify a motion primitive for each action, e.g., a trivial linear interpolation between starting and goal position. In any case, we focus only on \emph{task-level} learning and refinement, thus motion learning is out of the scope of this paper.}.

We start from learned axioms with $A_{sem}$.
In fact, the methodology explained in Section \ref{sec:met-refine} is not affected by the specific action identification algorithm.

We execute the autonomous framework in the three ordered scenarios depicted in Figure \ref{fig:sims}.
The scene in Figure \ref{subfig:sim1} shows a specific task situation with rings to be placed on pegs without transfer. Transferring is however required for rings in Figure \ref{subfig:sim2}. Finally, Figure \ref{subfig:sim3} depicts a slightly more challenging scenario with occupied pegs. Thus, the sequence of selected scenarios, while relatively short (42 total actions, fewer than the training dataset used in Section \ref{sec:res-learn}), requires progressive task knowledge refinement and encodes main variants of peg transfer.

Table \ref{table:results-refine} shows the quantitative analysis on the rules learned after the three simulated scenarios were completed (the final task knowledge is reported in Appendix \ref{sec:online}).
Bold entries indicate an improved F1-score at the end of execution, meaning that learned axioms are more accurate.
In general, the average F1-score increases from $88\%$ to $93\%$, i.e., overall task knowledge is refined. 
Axioms for actions improve or maintain the same performance achieved from offline learning.
For effects of actions, \stt{at(A, ring, C)} achieves lower performance. 
A closer look at finally learned axioms for the corresponding initiating conditions, in equations \eqref{eq:init_at_1}-\eqref{eq:init_at_2}, the atoms starts to be ground when either \stt{move(A, ring, C)} action is executed, or the arm which cannot reach the peg moves to center for transfer (in this case, in fact, the other arm reaches the ring at the transfer location).
These initiating conditions are complete for the atom, and extend the only partial knowledge encoded in the original axioms learned offline (Appendix \ref{app:offline_meli}).
However, the terminating conditions in equations \eqref{eq:term_at_1}-\eqref{eq:term_at_2} only respectively specify that \stt{at(A, ring, C)} stops holding when, at transfer location, the other arm moves to the peg to place the ring, or the ring is released on a peg. Compared with ground-truth axioms in equations \eqref{eq:term_at_orig_1}-\eqref{eq:term_at_orig_2}, these terminating conditions are still incomplete, since \stt{at(A, ring, C)} should stop holding \emph{whenever \stt{A} moves away}. Collecting examples for all such occurrencies would require more task iterations and variants. Moreover, these considerations hold for all atoms representing the current location of arms, thus justifying the slightly worse performance achieved for \stt{at(A, peg, C)} (higher IQR). The median improvement on \stt{at(A, center)}, instead, rises from the scarce examples for \stt{move(A, center, C)} in the dataset for offline learning (only 4 instances in scenarios depicted in Figure \ref{fig:execs}). Thus, while the results for the action do not vary after online ASP refinement, the results for its effect improve.

ILASP was called 13 times, i.e., $\approx 30\%$ times over all required actions to complete the task instances.
We remark that each time ILASP was invoked, the user voted the autonomous-generated action as invalid, hence not correct and useful (in the best case) or \emph{potentially unsafe}.
This demonstrates the fundamental role of our knowledge refinement architecture for autonomy implementation.

From a computational perspective, the median - IQR required time for running ILASP was \SI{2.81}{s} - \SI{6.07}{s}\footnote{The search space is the same as for offline learning.}, thus compatible with practical application of our framework to real robotic scenarios with non-strict real-time requirements. Moreover, the computational performance may be optimized by using the more efficient FastLAS extension to ILASP by \cite{law2020fastlas}. However, this aspect is out of the scope of this paper, which is more methodological.

\begin{table}[t]
    \caption{Quantitative results (median - IQR range) after task knowledge refinement in simulation. We aggregate \stt{initiated} and \stt{terminated} conditions for effects of actions. In brackets, we mark the atoms for which performance is invariant with ``$=$'', and the ones with better performance with ``$+$''}
    \label{table:results-refine}
    \begin{tabular*}{\hsize}{@{\extracolsep{\fill}}c|c|c|c@{}}
        \toprule
        Head & $Pr$ & $Rec$ & $F1$\\
        \midrule
        \stt{move(A, ring, C)} & $0.50 - 1.00$ & $1.00 - 0.00$ & $0.67 - 1.00$\\
        \midrule
        \stt{move(A, peg, C)} & $1.00 - 1.00$ & $1.00 - 0.00$ & $\mathbf{0.93 - 1.00} (+)$\\
        \midrule
        \stt{move(A, center, C)} & $1.00 - 1.00$ & $1.00 - 0.00$ & $1.00 - 1.00 (=)$\\
        \midrule
        \stt{grasp(A, ring, C)} & $1.00 - 0.00$ & $1.00 - 0.50$ & $1.00 - 1.00$\\
        \midrule
        \stt{extract(A, ring, C)} & $1.00 - 0.00$ & $1.00 - 0.00$ & $1.00 - 0.00 (=)$\\
        \midrule
        \stt{release(A)} & $1.00 - 0.00$ & $1.00 - 0.00$ & $\mathbf{1.00 - 0.00} (+)$\\
        \midrule
        \stt{closed\_gripper(A)} & $1.00 - 0.00$ & $1.00 - 0.00$ & $1.00 - 0.00 (=)$\\  
        \midrule
        \stt{at(A, center)}  & $1.00 - 0.00$ & $0.83 - 1.00$ & $\mathbf{0.90 - 1.00} (+)$\\  
        \midrule
        \stt{at(A, ring, C)} & $1.00 - 0.33$ & $0.63 - 0.89$ & $0.29 - 1.00 (=)$\\  
        \midrule
        \stt{at(A, peg, C)} & $1.00 - 0.00$ & $0.75 - 1.00$ & $0.83 - 1.00$\\  
        \midrule
        \stt{placed(ring, C1, peg, C2)} & $1.00 - 0.00$ & $1.00 - 0.00$ & $1.00 - 0.00 (=)$\\  
        \midrule
        \textbf{Mean} & & & $\mathbf{0.93} (+)$\\
        \bottomrule
    \end{tabular*}
\end{table}

\section{Discussion}\label{sec:disc}
Our experiments evidence the efficacy of our methodology for offline robotic task knowledge learning from raw data, and interactive online refinement via human feedback.

Thanks to the generalization capabilities of logical task formalization, offline learning via ILP is highly \underline{\emph{data- and time-efficient}}, and \underline{\emph{robust to non-homogeneous small datasets of executions}} (i.e., non-repetitive action sequences under diverse environmental conditions), \underline{\emph{regardless of the specific action identification algorithm}}.
Moreover, our framework for \underline{\emph{online incremental learning from expert feedback}} is time-efficient, allowing to easily refine task knowledge online, thanks to the \underline{\emph{interpretable interface}} provided by ASP formalism to the human expert. Since the human can forbid unwanted actions and a deterministic logical task theory is progressively refined, the overall task execution \underline{\emph{is guaranteed to be safe}}, neglecting human fallacies.

This work opens to several research problems and opportunities for improvement, as outlined in the following.

\paragraph{Prior knowledge of actions and environmental features}
We make the assumption that all possible task actions and environmental features, i.e., the ASP signature of the task, are known in advance (though the environmental features are almost task-agnostic, since they only describe relative positions of entities of the task).
In more complex robotic scenarios, involving, e.g., multiple robots with different capabilities (as in industrial robotics or real clinical setups), this may be infeasible.

A possible solution is to learn relevant task concepts from available textbooks, if available.
For instance, in the context of robotic surgery, pre-trained transformers have been employed by \cite{bombieri2023machine,bombieri2024surgicberta} to automatically retrieve relevant surgical concepts from natural language.
However, real robotic scenarios may still present novel unseen flows of execution, where the system must be able to learn about new actions and new environmental features. 
For such cases, we plan to extend our framework to support new concept learning, which requires some modifications to the current software architecture.

\paragraph{Advanced symbol grounding}
Given the complete signature of the ASP task, we rely on standard image and kinematic segmentation techniques to retrieve grounded symbols, following Rules \eqref{eq:ground_fluents}.
In more complex tasks with multiple sensor inputs, more sophisticated situation awareness may be required.

To this aim, in the future we plan to investigate the use of advanced symbol grounding techniques \citep{topan2021techniques}, recently applied to the problem of DRL under temporal logic specifications \citep{umili2023grounding}.

\paragraph{Task and motion learning}
An impelling requirement from the robotic research community is to jointly learn high-level routines and low-level motion strategies \citep{james2020rlbench}.
This can be achieved with hierarchical learning architectures \citep{sridharan2019reba}, or exploiting curriculum-based human demonstrations \citep{huang2024safety}.
However, these cannot be applied to hazardous scenarios as surgical robotics, where fast, safe and efficient learning is required from few examples.
For this reason, we will integrate online one-shot imitation learning with dynamic movement primitives \citep{ginesi2019dynamic,Ginesi2021}, to work simultaneously with the task-level ILP pipeline.

\paragraph{Bottom-up vs. top-down learning}
Our approach to task knowledge learning is bottom-up, i.e., it starts from raw data of robotic executions. Another approach to retrieve domain knowledge involves natural language processing applied to manuals and expert annotations, as shown, e.g., by \cite{appint,pan2023data,fuggitti2023nl2ltl}. 
In the future, we plan to combine the two methodologies, exploiting background knowledge extracted from domain-specific texts in the offline and online learning processes.

\paragraph{Probabilistic task knowledge}
At the moment we learn deterministic task specifications, accounting for uncertainty only when defining examples and covering the most of them accordingly (see Defintion \ref{def:ILASP_CDPI_noise}).
However, as shown, e.g., by \cite{berthet2016hubot}, a more accurate task representation may be provided by probabilistic logical specifications.
Learning probabilistic logical theories has been already explored, e.g., in the event calculus formalism by \cite{katzouris2020woled}. Alternatively, statistical relational learning by \cite{hazra2023deep,marra2024statistical} retrieves probabilistic knowledge graphs, but presents significant computational limitations for its applicability to complex robotic scenarios and online learning. 

In the future, we plan to investigate more sophisticated neurosymbolic planning and learning integrations accounting for task stochasticity.
One option is to exploit learned logical specifications for the efficient guidance of a probabilistically optimal planner \citep{meli2024learning}.
Alternatively, we could exploit our recent results in neurosymbolic DRL, to simultaneously learn a stochastic DRL policy enhanced by online refined ASP heuristics \citep{veronese2024online}. In terms of safety guarantees on the overall execution (currently yielded from the deterministic reasoning approach under expert supervision), they shall be then provided via stochastic formal verification \citep{marzari2024enumerating} or shielding techniques \citep{mazzi2023risk}. 

\paragraph{User validation}
Our work represents a preliminary architecture for interpretable human-robot interaction, where the human plays the role of a teacher and supervisor for the student robotic system.
Application to more complex tasks and real scenarios with physical robots will require an extended user study to assess the actual benefits of human-robot inter-play \citep{fosch2023role}, and the implementation of more sophisticated and user-oriented communication interfaces \citep{long2023human}, e.g., implementing voice commands with automatic speech-to-text and text-to-speech translation in the online refinement phase.

\section{Conclusion}\label{sec:conc}
In this paper, we have proposed a methodology based on ILP and ASP to discover task specifications, encoding action preconditions, constraints and effects, from raw video-kinematic datasets of robotic executions.
Our approach leverages on \emph{any} unsupervised action identification algorithm to retrieve ILP examples from recordings, combining them with almost task-agnostic commonsense concepts defined by an user, in order to increase the interpretability, trustability and social acceptance of learned axioms.

In the context of the peg transfer task with dual-arm dVRK, a benchmark training exercise for novice surgeons, we show that our offline learning pipeline requires very basic commonsense concepts, which are often task-agnostic since they only incorporate information about the location of robotic arms with respect to the environment, and the kinematic state of the grippers.
Moreover, the online learning algorithm is able to refine task knowledge even with few task examples.

This work represents an important step towards interpretable robotic task learning and generalization from limited raw recordings of execution, showing promising results for the scalability of our methodology to more complex and challenging robotic scenarios.
\newpage







\begin{appendices}

\section{Learned axioms}\label{sec:app}
\subsection{Unsupervised learning of ASP task knowledge}\label{sec:offline}
\subsubsection{$A_{sem}$}\label{app:offline_meli}
\begin{align*}
    &\stt{release(V1,t) :- not reachable(V1,peg,V2,t),not reachable(V1,ring,V2),}\\
    &\ \ \ \ \stt{closed\_gripper(V1,t),placed(ring,V2,peg,V3,t).}\\
    &\stt{grasp(V1,ring,V,t) :- not closed\_gripper(V1,t),at(V1,ring,V2,t).}\\
    &\stt{move(V1,peg,V2,t) :- at(V1,ring,V2,t),reachable(V1,peg,V2,t).}\\
    &\stt{move(V1,center,V2,t) :- not closed\_gripper(V3,t),at(V1,ring,V2,t),}\\
    &\ \ \ \ \stt{reachable(V3,peg,V2,t),V3!=V1.}\\
    &\stt{extract(V1,ring,V2,t) :- closed\_gripper(V1,t),placed(ring,V2,peg,V3,t),}\\
    &\ \ \ \ \stt{at(V4,ring,V2,t).}\\
    &\stt{move(V1,ring,V2,t) :- not closed\_gripper(V1,t),reachable(V1,ring,V2,t),}\\
    &\ \ \ \ \stt{reachable(V1,peg,V2,t).}\\
    &\stt{terminated(at(V1,peg,V2),t+1) :- move(V1,center,V3,t),color(V2).}\\
    &\stt{initiated(at(A,peg,C),t+1) :- move(A,peg,C,t).}\\
    &\stt{terminated(at(V1,ring,V2),t+1) :- move(V1,peg,V3,t),color(V2).}\\
    &\stt{initiated(at(V1,ring,V2),t+1) :- move(V1,ring,V2,t).}\\
    &\stt{terminated(closed\_gripper(A),t+1) :- release(A,t).}\\
    &\stt{initiated(closed\_gripper(A),t+1) :- grasp(A,ring,C,t).}\\
    &\stt{terminated(placed(ring,C1,peg,C2),t+1) :- extract(A,ring,C1,t),color(C2).}\\
    &\stt{initiated(placed(ring,V3,peg,V1),t+1) :- move(V2,peg,V1,t),color(V3).}\\
    &\stt{initiated(at(V1,center),t+1) :- move(V1,center,V2,t).}
\end{align*}

\subsubsection{$A_{kin}$}\label{app:offline_desp}
\begin{align*}
    &\stt{move(V1,ring,V2,t) :- arm(V1),color(V2).}\\
    &\stt{extract(V1,ring,V2,t):-not reachable(V4,peg,V2,t),closed\_gripper(V1,t),}\\ 
    &\ \ \ \ \stt{placed(ring,V2,peg,V3,t),at(V4,ring,V2,t).}\\
    &\stt{grasp(V1,ring,V2,t) :- at(V1,peg,V2,t),reachable(V1,ring,V2,t).}\\
    &\stt{grasp(V1,ring,V2,t) :- not closed\_gripper(V1,t),at(V1,ring,V2,t).}\\
    &\stt{release(V1,t) :- closed\_gripper(V1,t),at(V2,peg,V3,t).}\\
    &\stt{release(V1,t) :- placed(ring,V2,peg,V3,t),at(V1,ring,V3,t).}\\
    &\stt{release(V1,t) :- not reachable(V1,peg,V2,t),not reachable(V1,ring,V2,t),}\\ 
    &\ \ \ \ \stt{closed\_gripper(V1,t),placed(ring,V2,peg,V3,t).}\\
    &\stt{release(V1,t) :- closed\_gripper(V1,t),placed(ring,V2,peg,V3,t),}\\
    &\ \ \ \ \stt{at(V4,center,t),at(V1,ring,V2,t).}\\
    &\stt{move(V1,peg,V2,t):-at(V1,ring,V2,t),reachable(V1,peg,V2,t).}\\
    &\stt{move(V1,center,V2,t) :- not closed\_gripper(V3,t),at(V1,peg,V4,t),}\\
    &\ \ \ \ \stt{reachable(V3,peg,V2,t),V3!=V1.}\\
    &\stt{move(V1,center,V2,t) :- not closed\_gripper(V3,t),at(V1,ring,V2,t),}\\
    &\ \ \ \ \stt{reachable(V3,peg,V2,t),V1!=V3.}\\
    &\stt{move(V1,center,V2,t) :- not closed\_gripper(V1,t),at(V3,center,t),}\\
    &\ \ \ \ \stt{reachable(V1,peg,V2,t),reachable(V1,ring,V2,t),V1!=V3.}\\
    &\stt{terminated(at(V1,peg,V2),t+1) :- move(V1,center,V3,t),color(V2).}\\
    &\stt{initiated(at(A,peg,C),t+1) :- move(A,peg,C,t).}\\
    &\stt{terminated(at(V1,ring,V2),t+1) :- move(V1,peg,V3,t),color(V2).}\\
    &\stt{terminated(at(V1,ring,V2),t+1) :- release(V1,t),color(V2).}\\
    &\stt{initiated(at(V1,ring,V2),t+1) :- move(V1,center,V2,t).}\\
    &\stt{terminated(closed\_gripper(A),t+1) :- release(A,t).}\\
    &\stt{initiated(closed\_gripper(A),t+1) :- grasp(A,ring,C,t).}\\
    &\stt{initiated(placed(ring,V3,peg,V1),t+1) :- release(V2,t),color(V3),}\\
    &\ \ \ \ \stt{color(V1).}\\
    &\stt{initiated(at(V1,center),t+1) :- move(V1,center,V2,t).}
\end{align*}

\subsubsection{Baseline algorithm}
\begin{align*}
    &\stt{extract(V1,ring,V2,t) :- at(V1,ring,V2,t).}\\
    &\stt{grasp(V1,ring,V2,t) :- at(V1,ring,V2,t).}\\
    &\stt{release(V1,t) :- at(V1,ring,V2,t).}\\
    &\stt{release(V1,t) :- not reachable(V1,peg,V2,t),closed\_gripper(V1,t),}\\
    &\ \ \ \ \stt{placed(ring,V2,peg,V3,t).}\\
    &\stt{release(V1,t) :- not closed\_gripper(V2,t),reachable(V1,peg,V3,t),}\\
    &\ \ \ \ \stt{reachable(V2,ring,V3,t).}\\
    &\stt{move(V1,peg,V2,t) :- closed\_gripper(V1,t),color(V2).}\\
    &\stt{move(V1,ring,V3,t) :- placed(ring,V2,peg,V3,t),reachable(V1,ring,V3,t).}\\
    &\stt{move(V1,ring,V2,t) :- at(V3,center,t),reachable(V1,ring,V2,t).}\\
    &\stt{move(V1,ring,V2,t) :- not at(V1,center,t),not reachable(V1,peg,V2,t),}\\
    &\ \ \ \ \stt{reachable(V1,ring,V2,t).}\\
    &\stt{move(V1,center,V2,t) :- closed\_gripper(V1,t),reachable(V4,peg,V2,t),V4!=V1.}\\
    &\stt{terminated(at(V1,peg,V2),t+1) :- move(V1,center,V3,t),color(V2).}\\
    &\stt{initiated(at(A,peg,C),t+1) :- move(A,peg,C,t).}\\
    &\stt{terminated(at(V1,ring,V2),t+1) :- move(V1,peg,V3,t),color(V2).}\\
    &\stt{initiated(at(V1,ring,V2),t+1) :- move(V1,ring,V2,t).}\\
    &\stt{terminated(closed\_gripper(A),t+1) :- release(A,t).}\\
    &\stt{initiated(closed\_gripper(A),t+1) :- grasp(A,ring,C,t).}\\
    &\stt{terminated(placed(ring,C1,peg,C2),t+1) :- extract(A,ring,C1,t),color(C2).}\\
    &\stt{initiated(placed(ring,V3,peg,V1),t+1) :- move(V2,peg,V1,t),color(V3).}\\
    &\stt{initiated(at(V1,center),t+1) :- move(V1,center,V2,t).}
\end{align*}

\subsection{Online supervised refinement of ASP task knowledge}\label{sec:online}
\begin{subequations}
\begin{align}
    &\stt{release(V1,t) :- closed\_gripper(V1,t).}\\
    &\stt{grasp(V1,ring,V2,t) :- not closed\_gripper(V1,t),at(V1,ring,V2,t).}\\
    \nonumber&\stt{move(V1,peg,V2,t) :- not at(V1,peg,V2,t),closed\_gripper(V1,t),}\\
    &\ \ \ \ \stt{reachable(V1,peg,V2),not at(V3,center),arm(V3).}\\
    \nonumber&\stt{move(V1,peg,V2,t) :- not at(V1,peg,V2,t),closed\_gripper(V1,t),}\\
    &\ \ \ \ \stt{reachable(V1,peg,V2),at(V3,center),not closed\_gripper(V3).}\\
    &\stt{move(V1,center,V2,t) :- at(V1,ring,V2,t).}\\
    \nonumber&\stt{extract(V1,ring,V2,t) :- closed\_gripper(V1,t),placed(ring,V2,peg,V3,t),}\\
    &\ \ \ \ \stt{at(V4,ring,V2,t).}\\
    &\stt{move(V1,ring,V2,t) :- reachable(V1,ring,V2).}\\
    \nonumber&\stt{terminated(at(V1,peg,V2),t+1) :- placed(ring,V3,peg,V3,t),}\\
    &\ \ \ \ \stt{reachable(V1,peg,V2),move(V1,ring,V4,t).}\\
    \nonumber&\stt{initiated(at(A,peg,C),t+1) :- move(A,peg,C,t).}\\
    \label{eq:term_at_1}&\stt{terminated(at(V1,ring,V2),t+1) :- at(V1,center,t),move(V3,peg,V2,t).}\\
    \nonumber&\stt{terminated(at(V1,ring,V2),t+1) :- release(V1,t),reachable(V1,ring,V2),}\\
    \label{eq:term_at_2}&\ \ \ \ \stt{at(V1,peg,V3,t).}\\
    \label{eq:init_at_1}&\stt{initiated(at(V1,ring,V2),t+1) :- move(V1,ring,V2,t).}\\
    \label{eq:init_at_2}&\stt{initiated(at(V1,ring,V2),t+1) :- reachable(V1,peg,V2),move(V3,center,V2,t).}\\
    &\stt{terminated(closed\_gripper(A),t+1) :- release(A,t).}\\
    &\stt{initiated(closed\_gripper(A),t+1) :- grasp(A,ring,C,t).}\\
    \nonumber&\stt{terminated(placed(ring,V1,peg,V2),t+1) :- reachable(V3,peg,V2),}\\
    &\ \ \ \ \stt{extract(V3,ring,V1,t),at(V3,ring,V1,t).}\\
    \nonumber&\stt{initiated(placed(ring,V1,peg,V2),t+1) :- release(V3,t),at(V3,ring,V1,t),}\\
    &\ \ \ \ \stt{at(V3,peg,V2,t).}\\
    &\stt{terminated(at(V1,center),t+1) :- reachable(V1,peg,V2),move(V3,center,V2,t).}\\
    &\stt{initiated(at(V1,center),t+1) :- move(V1,center,V2,t).}
\end{align}
\end{subequations}

\end{appendices}

\section*{Declarations}
\textbf{Competing interests} - The authors have no competing interests to declare that are relevant to the content of this article.\\
\textbf{Funding} - This project has received funding from the European Research Council (ERC) under the European Union's Horizon 2020 research and innovation programme under grant agreement No. 742671 (ARS).\\
\textbf{Ethics approval} - Not applicable.\\
\textbf{Consent to participate} - Not applicable.\\
\textbf{Consent for publication} - Not applicable.\\
\textbf{Availability of data and material} - The data and material are available at \href{https://github.com/danm11694/ILASP_robotics.git}{https://github.com/danm11694/ILASP\_robotics.git}\\
\textbf{Code availability} - The code is available at \href{https://github.com/danm11694/ILASP_robotics.git}{https://github.com/danm11694/ILASP\_robotics.git}\\
\textbf{Authors' contributions} - Daniele Meli contributed with the methodology, implementation, experiments and writing. Paolo Fiorini contributed with writing and supervision.

\bibliography{sn-bibliography}

\end{document}